\documentclass[twoside,leqno,twocolumn]{article}

\usepackage[letterpaper]{geometry}

\usepackage{ltexpprt}
\usepackage{hyperref}
\usepackage{complexity}
\usepackage[binary-units=true]{siunitx}
\usepackage{todonotes}
\usepackage{amsfonts}
\usepackage{nicefrac}
\usepackage{amsmath}
\usepackage[capitalize,]{cleveref}
\usepackage{booktabs}

\usepackage{caption}
\usepackage{subcaption}
\usepackage{graphicx}
\usepackage[algo2e]{algorithm2e}
\SetKwInput{KwInput}{Input}                
\SetKwInput{KwOutput}{Output}              
\begin{document}

\newcommand\relatedversion{}
\renewcommand\relatedversion{\thanks{The full version of the paper can be accessed at \protect\url{https://arxiv.org/abs/1902.09310}}} 

\title{\Large Near-Optimal Coverage Path Planning with Turn Costs}
\author{Dominik Krupke\thanks{Department of Computer Science, TU Braunschweig, 38106 Braunschweig, Germany, krupke@ibr.cs.tu-bs.de}}

\date{}

\maketitle







\begin{abstract} \small\baselineskip=9pt 
  
Coverage path planning is a fundamental challenge in robotics, with diverse applications in aerial surveillance, manufacturing, cleaning, inspection, agriculture, and more.
The main objective is to devise a trajectory for an agent that efficiently covers a given area,
while minimizing time or energy consumption.
Existing practical approaches often lack a solid theoretical foundation, relying on purely heuristic methods,
or overly abstracting the problem to a simple Traveling Salesman Problem in Grid Graphs.
Moreover, the considered cost functions only rarely consider turn cost, prize-collecting variants for uneven cover demand, or arbitrary geometric regions.


In this paper, we describe an array of systematic methods for handling arbitrary meshes derived from intricate, polygonal environments.
This adaptation paves the way to compute efficient coverage paths with a robust theoretical foundation for real-world robotic applications.
Through comprehensive evaluations, we demonstrate that the algorithm also exhibits low optimality gaps, while efficiently handling complex environments. 
Furthermore, we showcase its versatility in handling partial coverage and accommodating heterogeneous passage costs, offering the flexibility to trade off coverage quality and time efficiency.

\end{abstract}

\section{Introduction}

Coverage path planning is an important problem for various applications such as aerial surveillance~\cite{cabreira2019survey}, cleaning~\cite{bormann2018indoor}, milling~\cite{marshall1994survey}, mowing~\cite{hameed2014intelligent}, pest control~\cite{drone_vid}, and more.
It has already received a considerable amount of attention, mostly from a practical perspective, but also with some theoretical results.
The problem is provably hard to solve on multiple levels, as it contains \NP- and \PSPACE-hard problems such as the \textsc{Traveling Salesman Problem} (TSP), \textsc{Covering}, and the \textsc{Piano Mover Problem}.

The simplest theoretical abstraction of the problem is the \textsc{TSP in Grid Graphs}.
Here, we simply place a grid, with a cell size matching the agent's coverage capabilities, over the area and compute the shortest tour on it.
Because the TSP appears in many applications, it is one of the most well researched optimization problems, such that there are highly capable solvers despite its proved hardness.
The Concorde solver~\cite{applegateconcorde} is able to solve instances with tens of thousands of vertices to proved optimality~\cite{applegate2009certification} and there are other algorithms that can compute good solutions for much larger instances.
Concorde is also used to optimize coverage paths, e.g., by Bormann et al.~\cite{bormann2018indoor}.

Although solving the TSP in Grid Graphs aims to minimize tour length, which is an important factor in energy consumption, this narrow optimization criterion can lead to unintended consequences.
In applications such as multicopters, straighter flight paths are generally more energy-efficient~\cite{cabreira2018energy, modares2017ub}.
An objective focused solely on minimizing the length of a coverage tour often encourages wavy routes, as this approach enables, e.g., covering two lanes in a single pass.
Consequently, these ostensibly shorter tours can actually be more expensive to execute.

This issue is addressed in the problem \textsc{Milling with Turn Costs}, which not only minimizes the length but also the sum of turn angles the tour performs through the grid~\cite{arkin2005optimal}.
While still not capturing all the dynamics, it serves as a more realistic approximation for various scenarios and mitigates the shortcomings of focusing solely on length minimization.
Unfortunately, turn costs increase the complexity of the problem such that not only itself but already the cycle cover relaxation becomes \NP-hard~\cite{CIAC2019}.
While the optimally solvable problem size increased from less than \num{100} vertices~\cite{de2011experimental} to over \num{1000} vertices~\cite{ALENEX19}, the still large difference to classical TSP shows the limits of computing optimal solutions for realistic dynamic models, even for strongly simplified environments.

Besides complex dynamics, we sometimes do not need to cover the whole area.
A true \SI{100}{\percent} coverage is in many cases even not achievable because the tool simply does not fit into every corner. 
Instead, we have a feasible area that allows us to move in, and a smaller subset of it that is actually `valuable'.
A vacuum robot can move within the whole room, but often there are dirt-prone areas and cleaner areas, which do not need to be cleaned every time.
A harvester can move along the whole field, but crop yield can be heterogeneous; the harvester does not need to harvest everything, rather only most of the harvest.
For aerial supervision, there are areas of higher and lower interest.
Additionally, there may be areas that are harder to pass than others, e.g., wind fields for UAVs~\cite{ware2016analysis} and difficult terrain or inclinations~\cite{hameed2014intelligent} for ground-based vehicles.

\begin{figure}[t]
  \begin{subfigure}[b]{0.49\columnwidth}
    \includegraphics[width=\columnwidth]{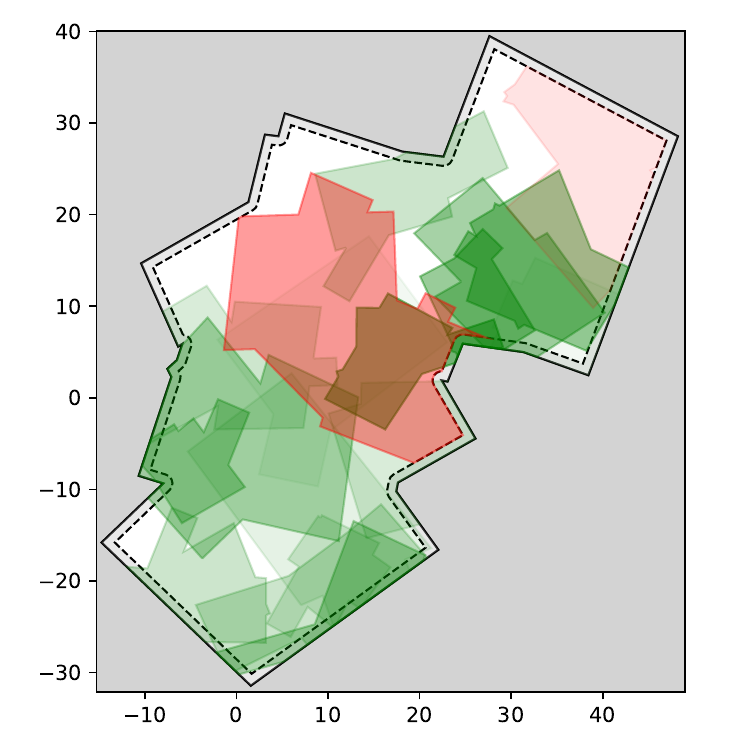}
    \caption{Polygonal instance.}
  \end{subfigure}
  \begin{subfigure}[b]{0.49\columnwidth}
    \includegraphics[width=\columnwidth]{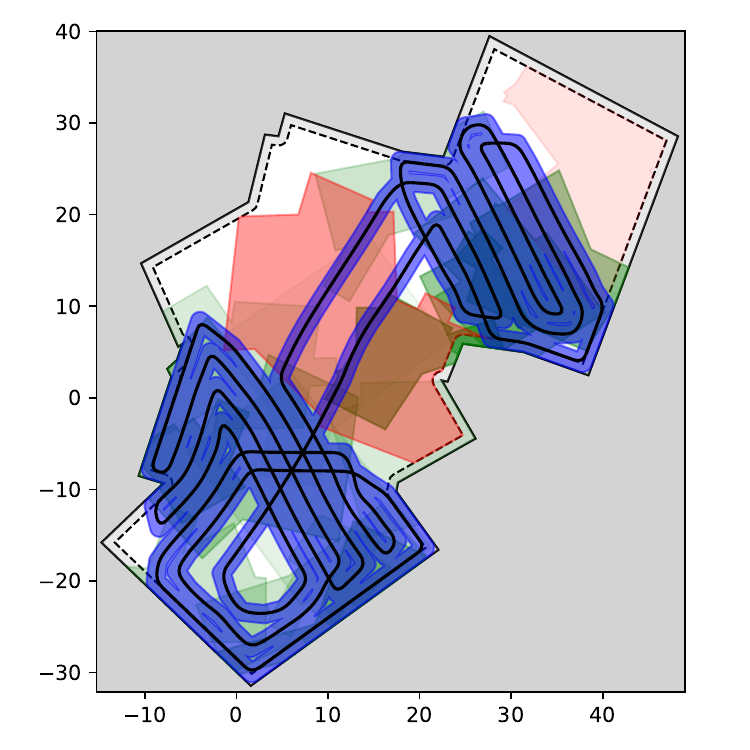}
    \caption{Solution after meshing.}
  \end{subfigure}
  \caption{
  A complex polygonal instance (a) is discretized using a meshing-algorithm in which the trajectory (b) is computed.
  Green indicates important areas, red indicates increased passage costs, and blue indicates the covered area of the black trajectory.
  We see that the trajectory minimizes the turn costs and focuses on the important areas while the expensive areas are avoided.
  }\label{fig:cpp:prac:example}
\end{figure}

Fekete and Krupke~\cite{CIAC2019,ALENEX19} proposed a constant-factor approximation algorithm for the \textsc{Milling with Turn Costs} problem on grid graphs, which is also able to handle partial coverage via skipping-penalties.
In this paper, we generalize this algorithm to work on arbitrary meshes obtained from polygonal environments and heterogeneous costs, which allows us to compute efficient trajectories based on a theoretical foundation for real-world applications, see \cref{fig:cpp:prac:example}.
We show in our evaluation that the algorithm is able to compute solutions that are on average close to optimum (\SIrange{10}{15}{\percent}), on the mesh representation.
While the constant-factor approximation guarantee may be lost for arbitrary meshes, this paper shows how a theoretical algorithm for coverage path planning on square grids can be generalized for real-world applications.

\subsection{Related Work}\label{sec:cpp:practical:related}


Planning a trajectory for a tool to cover an area, e.g., mowing a field or vacuuming a room, is known as the \emph{Coverage Path Planning} problem (CPP).
The CPP already enjoyed a lot of attention for different applications, models (e.g., multi-robot), constraints, and objectives, as can be seen in multiple surveys~\cite{choset2001coverage,galceran2013survey,bormann2018indoor,cabreira2019survey}.
There are multiple approaches, the two most prominent being: (1) decomposing the larger area into simpler areas that can be covered using spiraling or zigzag patterns~(\cite{oksanen2009coverage,choset1998coverage,choset2000exact}) and (2) applying a (regular square) grid onto the area,
 where each grid cell roughly represents the coverage area, converting the geometric coverage problem into a discrete touring problem on grid graphs~(\cite{bormann2015new,murtaza2013priority,zheng2010multirobot,sharma2019optimal,modares2017ub}).
In this paper, we use the second approach but generalized to arbitrary meshes that can adapt better to the area than strict grids, as a well fitting mesh can drastically improve the achievable tours.
When only considering the length of the trajectory, the problem becomes the famous \textsc{Traveling Salesman Problem} (TSP), which is NP-hard even in square grids~\cite{itai1982hamilton}, but can be solved well in practice due to extensive algorithm engineering~\cite{applegate2011traveling}.
To account for the non-negligible dynamics, we need to incorporate turn costs, which makes the problem significantly harder.
Even previously simple relaxations become \NP-hard~\cite{CIAC2019} but constant-factor approximations are available~\cite{arkin2001optimal,arkin2005optimal,CIAC2019}.
On grid graphs, instances with around \num{1000} vertices could be solved to optimality, and approximation algorithms have been applied to instances with up to \num{300000} vertices~\cite{ALENEX19}.
For general points in the plane, the problem is known as the \textsc{Angular Metric TSP}, and only a logarithmic approximation is known~\cite{aggarwal2000angular}.
A further generalization to abstract graphs is the \textsc{Quadratic TSP}, which plays an important role, e.g., in bioinformatics~\cite{fischer2014exact}.
Of these problems, only instances with less than \num{100} vertices can be expected to be solved to optimality in reasonable time~\cite{jager2008algorithms,rostami2013quadratic,oswin2017minimization}.
On the practical side, the CPP has been considered on models with distance and turn costs in various degrees,
such as only minimizing the number of turns~\cite{jensen2020near},
the sum of turn angles~\cite{bormann2015new,papachristos2016distributed,modares2017ub} (like this paper), 
or even model- and experiment-based cost functions~\cite{oksanen2009coverage,cabreira2018energy}.
Including heterogeneous cost functions have also been considered for CPP, e.g.,~\cite{zheng2010multirobot,hameed2014intelligent}, and for simple path planning~\cite{mitchell1991weighted,voros1999mobile,rowe2000finding}.

Another aspect of this paper is the ability to selectively cover the area, based on some value distribution.
There are a few paper that also consider partial coverage path planning.
Papachristos et al.~\cite{papachristos2016distributed} and Ellefsen, Lepikson, and Albiez~\cite{ellefsen2017multiobjective} consider partial inspection of three-dimensional structures with distance and turn costs.
Jensen et al.~\cite{jensen2020near} and Soltero et al.~\cite{soltero2014decentralized} perform coverage without a fixed radius, but minimize the distance of (weighted) points of interests to the trajectory.
Murtaza et al.~\cite{murtaza2013priority} compute a full-coverage of the area, but prioritize subareas based on a probability distribution to find targets quickly.
Sharma et al.~\cite{sharma2019optimal} also compute a full-coverage of the area, but with a limited budget, resulting in multiple tours that try to efficiently cover as much as possible.
However, all of these problems have significant differences to our problem.
On the theoretical side, there are the \emph{Penalty} and \emph{Budget TSP}, which allow skipping vertices at a penalty or try to cover as much as possible within a budget.
An overview of such problems is given by Ausiello et al.~\cite{ausiello2007prize}.


\subsection{Contributions}

In this paper, we make the following contributions:
\begin{itemize}
  \item We generalize an approximation algorithm for coverage tours in regular grid graphs to work on more realistic polygonal instances by using meshing algorithms, paving the way to compute more efficient coverage tours with a robust theoretical foundation for real-world applications.
  \item We approximate the dynamics of the agent by using a model based on a linear combination of the traveled distance and the sum of turn angles, as well as local multiplication factors for heterogeneous passage costs. This also allows creating soft obstacles, which should -- but do not have to -- be avoided.
  \item We investigate partial coverage by using a penalty for missed coverage, which allows trading off coverage quality and time efficiency. The area can be weighted to target important areas with the tour.
  \item We locally improve the tour by using a large neighborhood search (LNS), which is able to improve the tour by a few percent.
  \item We evaluate the optimality gap of the implementation on over \num{500} instances, which were semi-automatically generated to mimic real-world scenarios. Data and code are provided.
\end{itemize}

We do \emph{not} maintain the approximation factor of the original algorithm, but we show that the implementation is still able to compute good solutions on arbitrary meshes by using sound lower bounds.
Due to a lack of real-world instances and models, the evaluation is done only on synthetic instances, which were semi-automatically generated trying to mimic agricultural areas, locations with multiple buildings, and complex architecture.
A comparison to the geometric model without the restriction to a mesh is not performed, as strong lower bounds are difficult to obtain.
However, a comparison of the achievable solution quality of different grids and meshes was performed and the best meshing strategy was used for the evaluation.
We noticed that focusing on the coverage of the edges rather than on the coverage of the points improves results when choosing a mesh resolution.
Using hexagonal grids instead of square ones also shows beneficial, especially with higher turn costs.
Furthermore, it is important to note that not all meshing algorithms are well-suited for addressing our specific problem.
The corresponding study has been attached in \cref{sec:cpp:prac:grid}.

\subsection{Preliminaries}\label{sec:cpp:prac:prelim}

Given a graph $G=(P,E)$, where $P\subset \mathbb{R}^2$ is a set of waypoints, which span a potential trajectory, and $E$ are segments connecting two waypoints.
Additionally, we are given a value function $val: P \rightarrow \mathbb{R}^+$, which assigns a value to each waypoint, and a cost function $cost: P^3 \rightarrow \mathbb{R}^+$, which assigns a cost to each consecutive triple $u,v,w$ of waypoints with $uv, vw \in E$.
We call such a triple a \emph{passage} through the middle point $v$.
The goal is to find a tour $T=p_0, p_1,\ldots, p_{|T|-1}, p_0$ with $p_ip_{i+1}\in E$ for all $i\in \{0,\ldots, |T|-1\}$ that minimizes the objective
\begin{equation}
  \label{eq:cpp:obj}
  \min_T \underbrace{\sum_{i=0}^{{|T|-1}} cost(p_{i-1},p_{i},p_{i+1})}_{\text{Touring cost}} +\underbrace{\sum_{p\in P, p\not\in T} val(p)}_{\text{Coverage loss}} 
\end{equation}

We define the cost of using a passage $uvw$ by a linear combination of length of the two segments and their turn angle, weighted by $\tau\in \mathbb{R}^+$. 
It may additionally be scaled by a local factor $\alpha_v$. 
\[\text{cost}(u,v,w)=\alpha_v\cdot \left(\frac{d(u,v)+d(v,w)}{2} +\tau\cdot \text{turn}(u,v,w)\right)\]
The distance is halved to avoid double charging the edges.
See \cref{sec:cpp:prac:problemdef} for a more extensive discussion.
\section{Generalized Algorithm}\label{sec:cpp:prac:alg}

In this section, we show how to adapt the algorithm of Fekete and Krupke~\cite{CIAC2019,ALENEX19} to solve polygonal instances including expensive and valuable areas.
More precisely, we show how to approximate the area using an embedded graph, adapt the previous algorithm to work on arbitrary embedded graphs, and add optimizations.

The generalized algorithm has seven steps:
  (1) Convert the polygonal instances to a discrete graph of waypoints.
  (2) Compute a fractional solution in this graph using linear programming.
  (3) Select atomic strips using the fractional solution.
    This step is more complicated for general meshes than for square grids.
  (4) Perform a matching on the atomic strips and obtain a cycle cover.
  (5) Improve the cycle cover.
  (6) Connect the cycles to form a tour.
  (7) Improve the tour.
Steps (1), (3), (5),  and (7) significantly differ from the original algorithm, and we will describe them in detail.
However, if we are given a regular square grid for (1) and disable the local optimization, (5), and (7), the behavior of the algorithm is nearly identical to the original algorithm.

\begin{figure*}[htbp]
  \centering
  \includegraphics[width=0.8\textwidth]{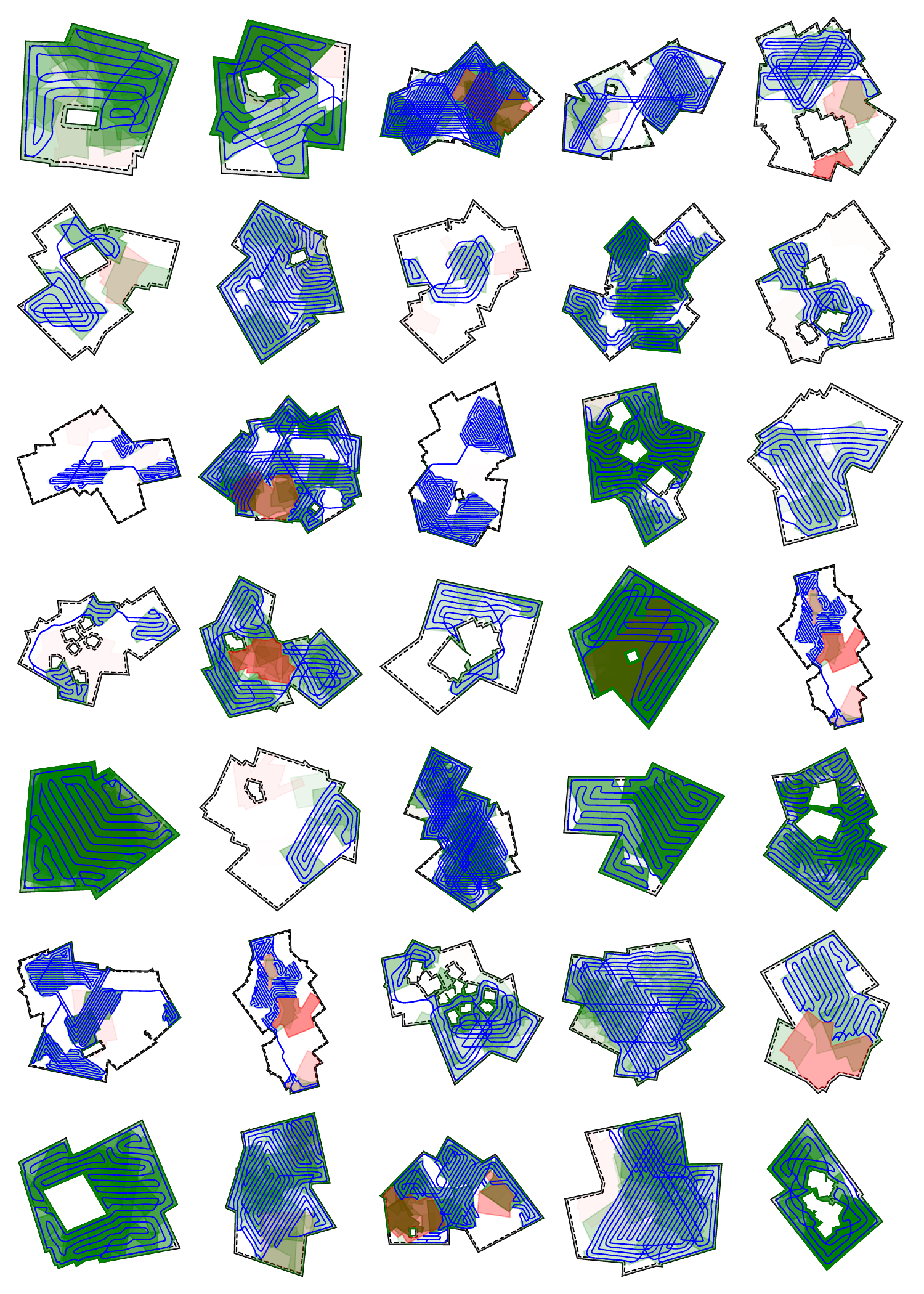}
  \caption{Examples of instances and solutions.
The weight for the turn costs vary, resulting in different tour characteristics (high turn costs lead to a higher redundancy and longer straight lines).
The trajectories are smoothed in post-processing using Bézier curves.
\vspace{-0.5cm}   
}\label{fig:cpp:prac:grid:partialexamples}
\end{figure*}

The resulting trajectories are shown in \cref{fig:cpp:prac:grid:partialexamples}.

\subsection[Discretization]{Step 1: Discretization}\label{sec:pcpp:alg:step1}

\begin{figure}[htb]
  \begin{subfigure}[b]{0.49\columnwidth}
    \includegraphics[width=\columnwidth]{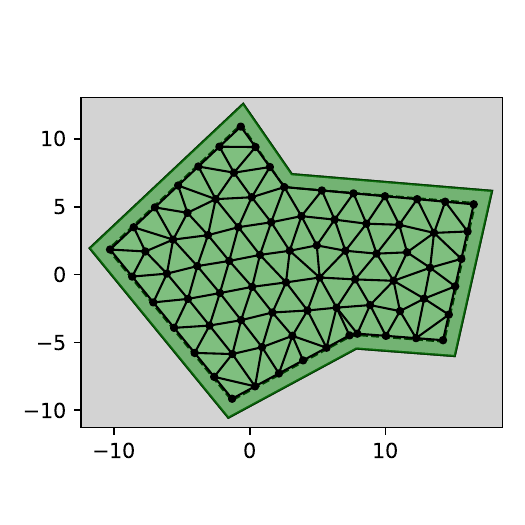}
    \caption{Meshing}\label{fig:mesh}
  \end{subfigure}
  \hfill
  \begin{subfigure}[b]{0.49\columnwidth}
    \includegraphics[width=\columnwidth]{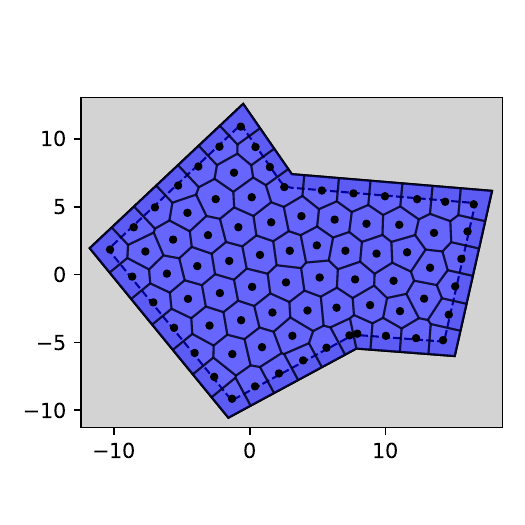}
    \caption{Voronoi cells.}\label{fig:voronoi}
  \end{subfigure}

  \caption{%
    To convert a polygonal instance into a graph, we first mesh the polygon (a) and use the coverage value of the Voronoi cells (b) to approximate the coverage value of each waypoint.
    }\label{fig:cpp:practical:conversion}
\end{figure}

We apply the meshing algorithm \emph{dmsh} (v0.2.17)\cite{nico_schlomer_2021_5019221} with additional smoothing by \emph{optimesh}~\cite{nico_schlomer_2021_4728056} onto the polygon to obtain a nicely fitting mesh, as can be seen in \cref{fig:mesh}.
The optimal distance between two waypoints is set to $0.95\cdot \nicefrac{4}{\sqrt{3}}\cdot r$, where $r$ represents the coverage radius, assuming the tool to have a circular coverage.
We set $r=1$ in the examples, but the algorithm works for any $r$.
The distance $\nicefrac{4}{\sqrt{3}}\cdot r\approx 3.31\cdot r$ in a triangular grid, which is approximated by the mesh, leads to parallel lines being a perfect $2\cdot r$ apart.
As \emph{dmsh} prefers vertices to be too far apart over too close, we counter this by reducing the distance by $\SI{5}{\percent}$.
Tours on this sparse grid will miss some area on turns, but we minimize turns and the missed coverage can be compensated by slightly enlarging the turns in post-processing.
The coverage value is estimated by the area covered by the Voronoi-cell of the waypoint, see \cref{fig:voronoi}.
We could also use the coverage of the agent at the waypoint, but this is less accurate because the coverage primarily happens when moving along the edges.
Getting a mesh that yields good tours is non-trivial, and a considerable number of experiments were necessary to find a good meshing algorithm and parameters.
Many aspects of the discretization also struggle with the infamous numeric issues of geometric operations, that have to be handled carefully.
Instead of \emph{dmsh} also the \emph{Packing of Parallelograms}-algorithm of \emph{gmsh}~\cite{geuzaine2009gmsh} can be used to obtain similar good meshes.
\emph{gmsh} is faster and more robust, but has more outliers regarding the quality than \emph{dmsh}.
There are many other meshing algorithms, but most of them are not suitable for our purpose as they will not allow smooth trajectories and equally sized cells but focus on different qualities.
More details can be found in \cref{sec:cpp:prac:grid}.

\subsection[Fractional Solution]{Step 2: Linear Relaxation}\label{sec:cpp:prac:alg:frac}


Given the graph $G=(P,E)$, we can obtain a fractional solution for a cycle cover by using linear programming.
We work on passages $uvw=wvu$ that cover a waypoint $v\in P$ coming from or going to the neighbored waypoints $u,w\in N(v)$.
For every passage $uvw$, the variable $x_{uvw}\geq 0$ denotes how often the passage is used.
Additionally, we use the variable $s_v\geq 0$ that denotes skipping the waypoint and paying for its coverage loss.

\begin{align}
  \min & \sum_{v\in P}\text{val}(v)\cdot s_v + \sum_{u, w \in N(v)} \text{cost}(u,v, w) \cdot x_{uvw} & \label{eq:cpp:prac:alg:frac:obj}
\end{align}
\begin{align}
  \text{s.t.}\quad & \sum_{u, w \in N(v)} x_{uvw} + s_v \geq 1 & \forall v \in P \label{eq:cpp:prac:alg:frac:covered}\\
  & 2\cdot x_{wvw} + \sum_{u\in N(v), u\not = w} x_{wvu} = \label{eq:cpp:prac:alg:frac:flow} \\
  & 2\cdot x_{vwv} + \sum_{u\in N(w), u\not = v} x_{vwu} & \forall vw \in E \nonumber
\end{align}

\begin{figure}
  \begin{subfigure}[b]{0.49\columnwidth}
  \includegraphics[width=\columnwidth]{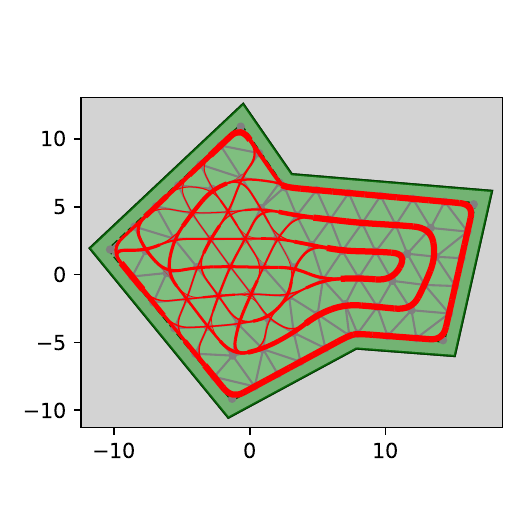}
  \caption{}\label{fig:cpp:prac:alg:fracsol1}
  \end{subfigure}
  \begin{subfigure}[b]{0.49\columnwidth}
  \includegraphics[width=\columnwidth]{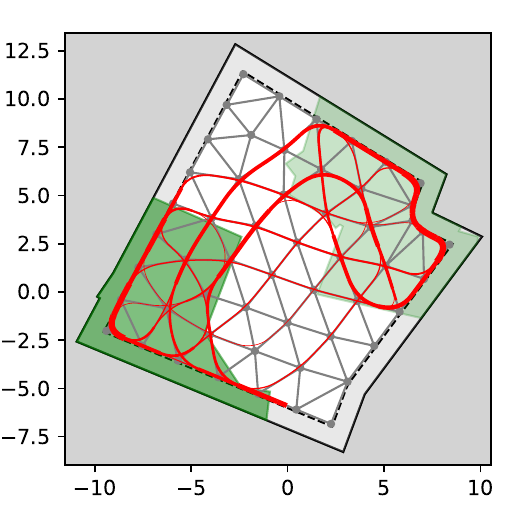}
  \caption{}\label{fig:cpp:prac:alg:fracsol2}
  \end{subfigure}
  \caption{
    Fractional solutions in red for full-coverage (a) and partial coverage (b). The thickness indicates the fractional values.
  }\label{fig:cpp:prac:alg:fracsol}
\end{figure}

The objective in \cref{eq:cpp:prac:alg:frac:obj} simply minimizes the missed coverage value and touring costs.
\Cref{eq:cpp:prac:alg:frac:covered} enforces a waypoint either to be covered or skipped, and \cref{eq:cpp:prac:alg:frac:flow} enforces a consistent flow, i.e., every edge is used equally from both sides.
Examples for fractional solutions covering the whole area or for partial coverage are given in \cref{fig:cpp:prac:alg:fracsol1} resp. \cref{fig:cpp:prac:alg:fracsol2}.

\subsection[Atomic Strips]{Step 3: Atomic Strips}\label{sec:pcpp:alg:step3}

In the next step, we want to compute a cycle cover, using the fractional solution of the previous solution as a hint.
If the costs would only depend on the distance, the cycle cover could efficiently be computed by a minimum-weight perfect matching.
For this, we would replace every waypoint by two vertices and connect them to all other vertices with the corresponding distance, efficiently calculable by Dijkstra's algorithm.
To implement partial coverage, we would add an edge with the corresponding value of the coverage loss between the two vertices of a waypoint.
The minimum-weight perfect matching would then either enforce every waypoint to have an incoming and an outgoing trajectory, i.e., be in a cycle, or only use the internal edge and skip the waypoint.

\begin{figure}[htb]
  \includegraphics[width=0.9\columnwidth]{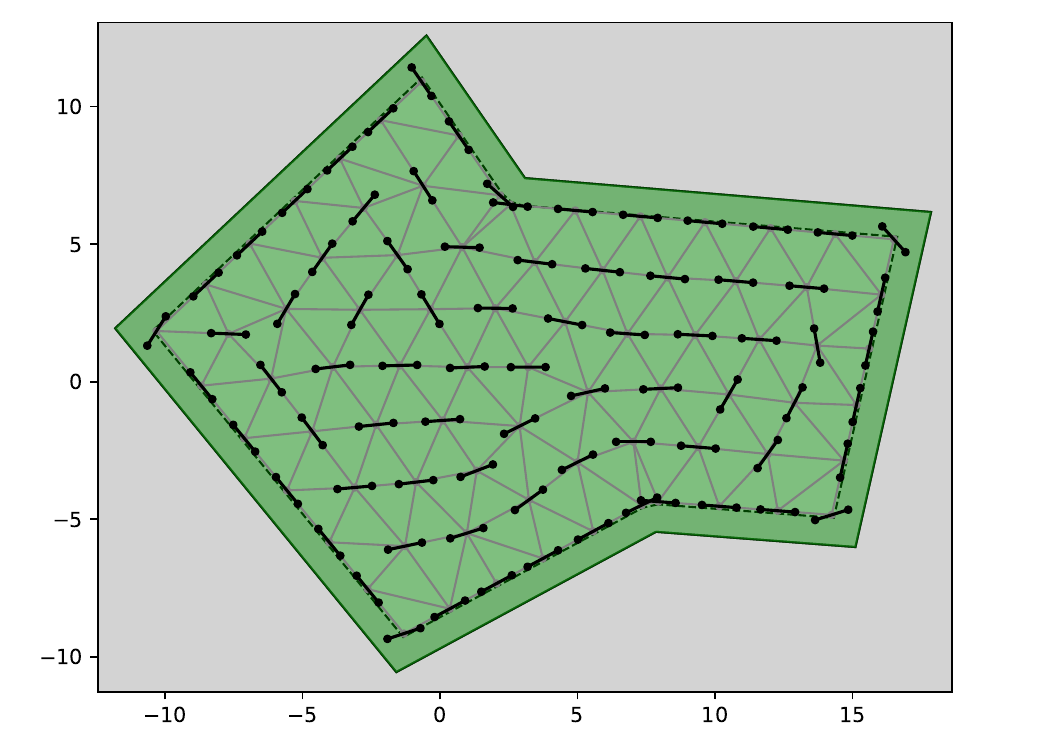}
  \caption{Replacing every waypoint by an atomic strip (black segments) converts the problem into a matching problem without losing the turn costs.
  The orientation of each atomic strip needs to be guessed from the fractional solution (\cref{fig:cpp:prac:alg:fracsol1}), and wrong guesses can degrade the solution.}\label{fig:cpp:prac:alg:atomicstripsexample}
\end{figure}

With turn costs, the cycle cover problem gets NP-hard, but Fekete and Krupke~\cite{CIAC2019} showed that we can use the fractional solution of the previous step
to estimate in which orientation  we go through a waypoint, and move the corresponding turn costs to the edge weights.
In square grids, this technique can be shown to yield a 4-approximation, and a 6-approximation in triangular grids.
This can be imagined as replacing every waypoint by an epsilon-length segment, as in \cref{fig:cpp:prac:alg:atomicstripsexample}, whose orientation is most used in the fractional solution.
We are calling these epsilon-length segments \emph{atomic strips}.
Computing a minimum-weight perfect matching on the endpoints, yields the optimal cycle cover that includes all these segments.
If the segments have been chosen correctly (which is NP-hard), the minimum-weight perfect matching actually corresponds to an optimal cycle cover on the waypoints.

Meshes make the selection of these atomic strips more complicated, as there can be more than just two or three sensible orientations.
A useful property of the atomic strips is that the larger the turn is, the more orientations are optimal.
For a U-turn, every orientation is optimal.
The straighter a passage, the more important a good orientation becomes; but often these cases are easy to guess from the fractional solution.
Therefore, it is sensible to limit the potential orientations to the orientations of incident edges, i.e., neighbors.
We weight each orientation by how well the passages of the fractional solution fit to it and chose the one with the highest sum.

\begin{figure}
  \centering
  \includegraphics[width=0.9\columnwidth]{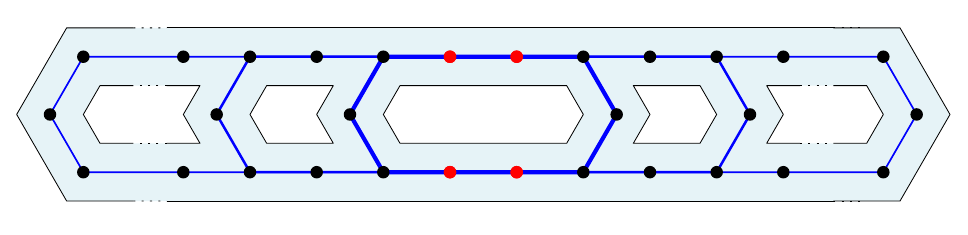}
  \caption{Optimal tours with turn costs in a regular triangular grid can require a linear amount of passages through some waypoints (red).}\label{fig:grid:hexaAtomicProblem}
\end{figure}

Connecting all waypoints with each other results in a quadratic number of edges, whose weights are non-trivial to compute.
Fekete and Krupke~\cite{ALENEX19} noted that it is more efficient to only connect the waypoints with their neighbors (also making the weights easy to compute), 
and to allow for optional atomic strips to deal with potentially necessary overlapping trajectories.
An optional atomic  strip can be implemented by simply adding an edge with zero weight between its endpoints, allowing it to be neutralized without additional costs.
Arkin et al.~\cite{arkin2005optimal} showed that in a square grid, every vertex is visited at most four time, limiting the number of necessary optional atomic strips.
For triangular grids, the number of necessary visitations can be linear, as shown in \cref{fig:grid:hexaAtomicProblem}, destroying the approximation factor when using this optimization.
However, this is an artificial instance, and in our instances, every waypoint is usually only covered once or twice.
A further  challenge is that the optional atomic strips also have to match the original trajectory of the longer edges to reconstruct the actual costs.
Otherwise, connecting two waypoints via optional atomic strips could be more expensive than connecting them directly.
Adding a number of optional strips for any orientation would solve this problem, but it would also increase the computational complexity.
Therefore, we limit the number of atomic strips per waypoint to a constant $k$, and additionally allowing every waypoint $p\in P$ at most one atomic strip per neighbor $n\in N(p)$.
This keeps the complexity of the auxiliary graph in $O(|P|\cdot k^2)$.

\begin{figure*}[htb]
  \begin{subfigure}[b]{0.24\textwidth}
    \includegraphics[width=\columnwidth]{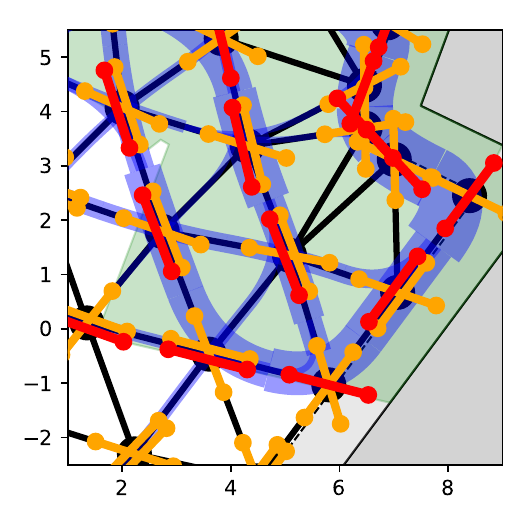}
    \caption{$k=3$}
  \end{subfigure}
  \begin{subfigure}[b]{0.24\textwidth}
    \includegraphics[width=\columnwidth]{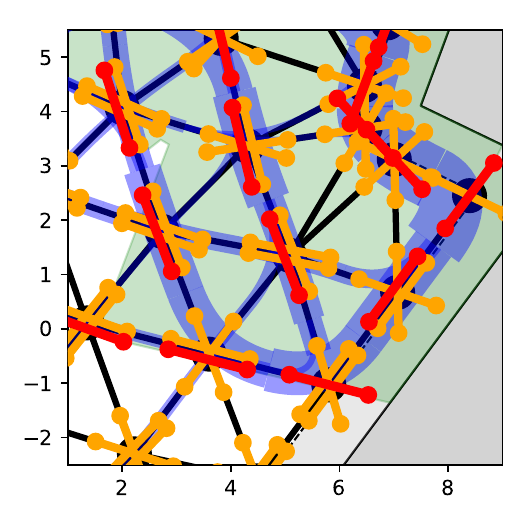}
    \caption{$k=4$}
  \end{subfigure}
  \begin{subfigure}[b]{0.24\textwidth}
    \includegraphics[width=\columnwidth]{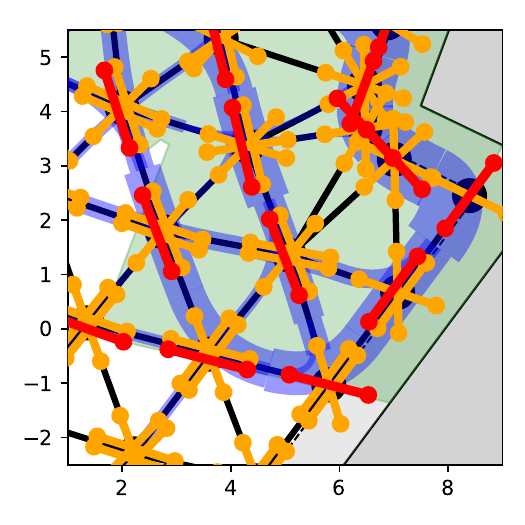}
    \caption{$k=5$}
  \end{subfigure}
  \begin{subfigure}[b]{0.24\textwidth}
    \includegraphics[width=\columnwidth]{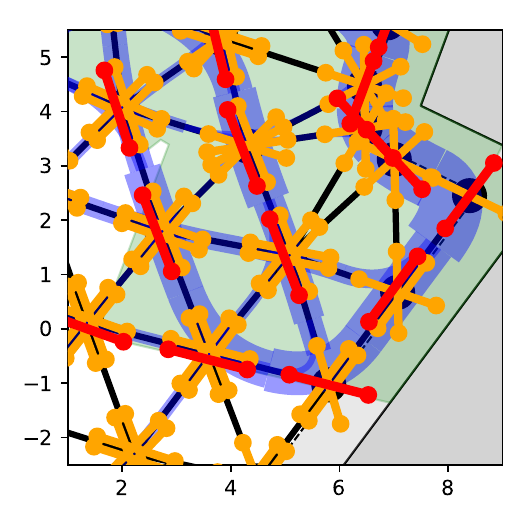}
    \caption{$k=6$}
  \end{subfigure}
  \caption{%
    Example of atomic strip selection for different $k$.
    The atomic strips are displayed in yellow (optional) and red (mandatory).
    The grid is displayed in black and the fractional solution in blue.
    }\label{fig:pcpp:alg:atomicstripsexample}
\end{figure*}

An example for different $k$ can be seen in \cref{fig:pcpp:alg:atomicstripsexample} and the detailed implementation is described in \cref{apdx:cpp:prac:step3}.

\subsection[Matching]{Step 4: Matching}

We are left with a weighted graph on the endpoints of the atomic strips, and we want to compute a minimal matching.
There are edges between any endpoints of atomic strips belonging to neighbored waypoints in the grid.
The weight corresponds to the touring costs between the two waypoints, with the corresponding orientation at the endpoints.
Additionally, each atomic strip has an edge between its two endpoints.
For the mandatory atomic strip, the weight corresponds to the opportunity loss, i.e., the assigned coverage value, when not covering it.
For all others, the cost is zero to allow skipping them without additional costs.
Let $k$ be the maximal number of atomic strips at a waypoint, then the number of vertices and edges in the matching instance is in $O(|P|\cdot k^2)$.

We solve the corresponding minimum-weight perfect matching instance with the Blossom~V algorithm of Kolmogorov~\cite{Kolmogorov2009}.
The author states a worst-case complexity of $O(n^3m)$, which would be prohibitive, but in practice it shows to be sufficiently fast even for large instances.
Connecting the atomic strips via the matched endpoints yields a set of cycles, see \cref{fig:cpp:prac:alg:matchingcycles}, that we can connect in Step 6.

\subsection[Local Optimization]{Step 5: Local Optimization}\label{sec:cpp:prac:alg:ccopt}

\begin{figure}[htb]
  \begin{subfigure}[b]{0.24\columnwidth}
    \includegraphics[width=\columnwidth]{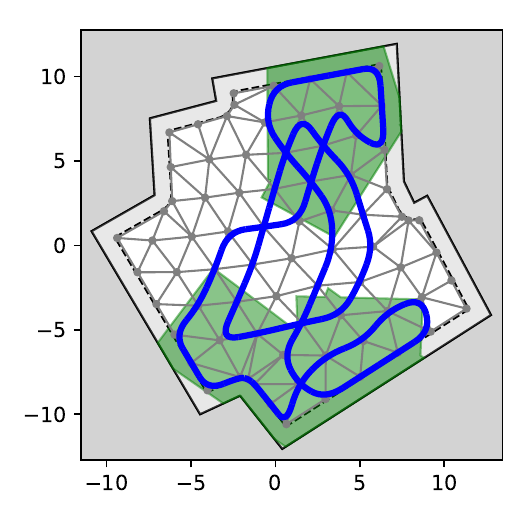}
    \caption{}\label{fig:cpp:prac:alg:ccopt:a}
  \end{subfigure}
  \begin{subfigure}[b]{0.24\columnwidth}
    \includegraphics[width=\columnwidth]{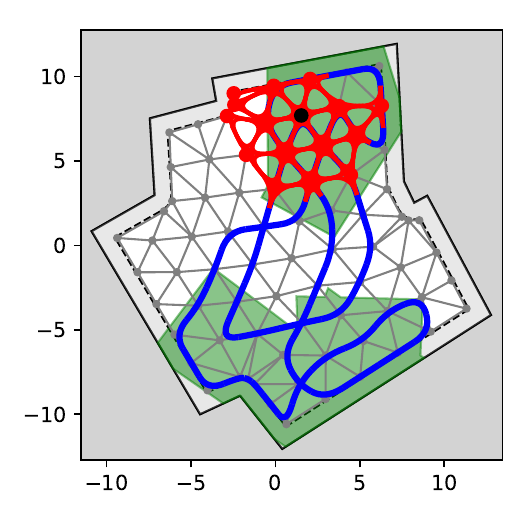}
    \caption{}\label{fig:cpp:prac:alg:ccopt:b}
  \end{subfigure}
  \begin{subfigure}[b]{0.24\columnwidth}
    \includegraphics[width=\columnwidth]{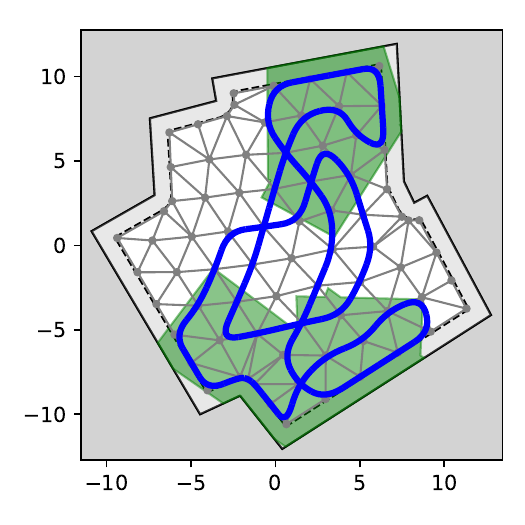}
    \caption{}\label{fig:cpp:prac:alg:ccopt:c}
  \end{subfigure}
  \begin{subfigure}[b]{0.24\columnwidth}
    \includegraphics[width=\columnwidth]{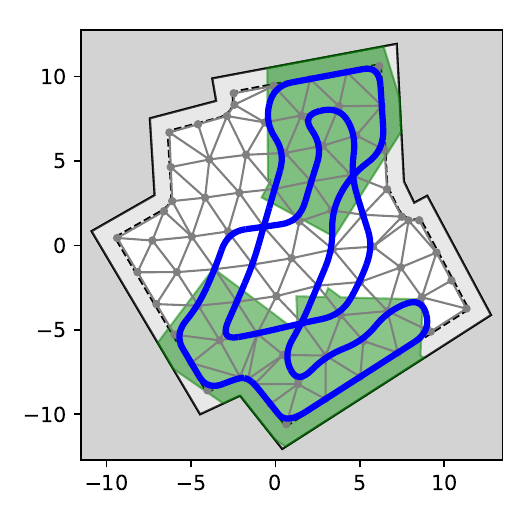}
    \caption{}\label{fig:cpp:prac:alg:ccopt:d}
  \end{subfigure}
  \caption{By optimizing local areas (red) of cycle covers (blue) with mixed integer programming, we can improve the initial cycle cover.
  The solution provided by the previous approach without optimizations is shown in \cref{fig:cpp:prac:alg:ccopt:a}.
  We then select an expensive area (\cref{fig:cpp:prac:alg:ccopt:b}) and optimize it to near optimality, resulting in (\cref{fig:cpp:prac:alg:ccopt:c}).
  After five such iterations, we end with a visibly improved solution (\cref{fig:cpp:prac:alg:ccopt:d}).
  By chance, the optimized solution is even connected.
  }\label{fig:cpp:prac:alg:ccopt}
\end{figure}

Before we continue to connect the cycles to a single tour, we can optimize the cycle cover.
For this, we select a small but expensive part of the solution and compute a (nearly) optimal solution via mixed integer programming.
This can be repeated multiple times until a satisfying solution is obtained, see \cref{fig:cpp:prac:alg:ccopt}.
Note that it is possible to solve many instances with \num{1000} vertices in regular square grids to optimality, as described in~\cite{icra_mosquito_2018,ALENEX19}.
Also, for irregular grids, small instances with less than \num{100} vertices can usually be solved within seconds.
We denote the desired number of vertices for local optimization by $t$.

We select the expensive area to be optimized by choosing an expensive root and selecting the first $t$ vertices of a breadth-first-search.
The expense of a waypoint in a solution is denoted as the cost of the passages covering it, or the corresponding opportunity loss if it is not used.
To make the selection more robust, we also include the expenses of all direct neighbors by summing them.

By simply replacing the fractional variables with integral variables, the linear program in \cref{sec:cpp:prac:alg:frac} yields a corresponding MIP\@.
In this MIP we fix all variables of the given solution except the variables corresponding to the $t+1$ selected waypoints.
Of course, we do not need to include the fixed waypoints in this MIP at all but only need to place the corresponding constants into \cref{eq:cpp:prac:alg:frac:flow}.
This ensures that the local solution remains consistent with the fixed exterior solution.
After optimizing the local MIP, we replace the part in the solution and exclude the root and its neighbors to be selected as root in further iterations.
This is necessary because the expensive parts can already be optimal (within their local area) and should not be optimized again.

A useful property of the MIP is that the optimization process usually is faster, if our (local) solution is already (nearly) optimal.
If we provide the MIP-solver with the corresponding start solution, it only has to find a matching lower bound.
Using the running time and the actual improvements, one could improve the selection of the next area, or dynamically increase it.
By choosing disjunct areas, this optimization approach also allows efficient parallelization.
However, we leave such optimizations to future work, and simply perform $i$ iterations for a fixed area size $t$.

\subsection[Connecting Cycles]{Step 6: Connecting Cycles}

\begin{figure}
  \begin{subfigure}[b]{0.49\columnwidth}
    \includegraphics[width=\columnwidth]{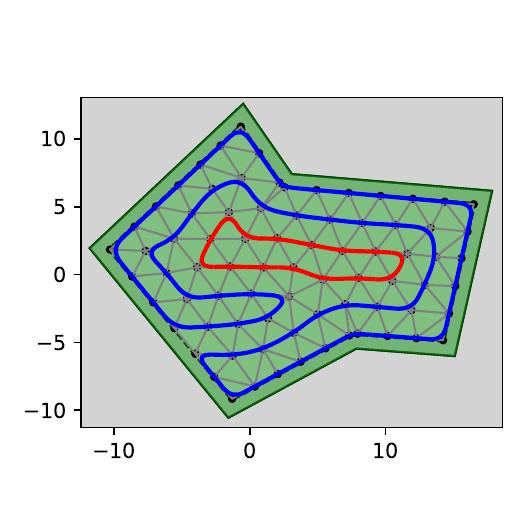}
    \caption{Cycle cover}\label{fig:cpp:prac:alg:matchingcycles}
  \end{subfigure}
  \begin{subfigure}[b]{0.49\columnwidth}
  \includegraphics[width=\columnwidth]{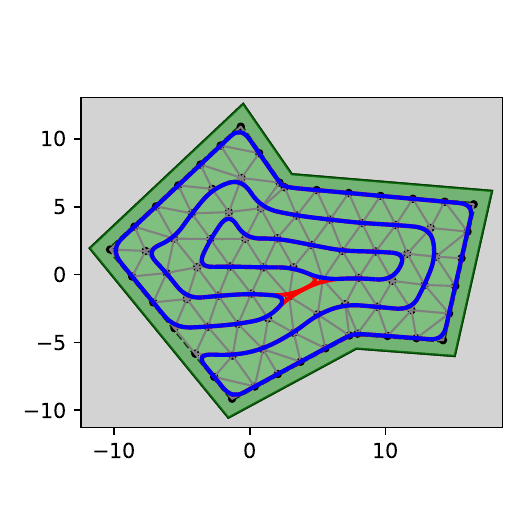}
  \caption{Connecting}\label{fig:cpp:prac:alg:connect1}
\end{subfigure}
\caption{The matching of the atomic strips of \cref{fig:cpp:prac:alg:atomicstripsexample} yields a set of tours (a).
In this case, a red and a blue cycle.
It can also directly decide not to cover some waypoints, but in this case the coverage values are very high.
This cycle cover is then connected to a tour (b) via and edge (red).
}
\end{figure}

\begin{figure}[htb]
  \begin{subfigure}[b]{0.49\columnwidth}
    \includegraphics[width=0.49\columnwidth]{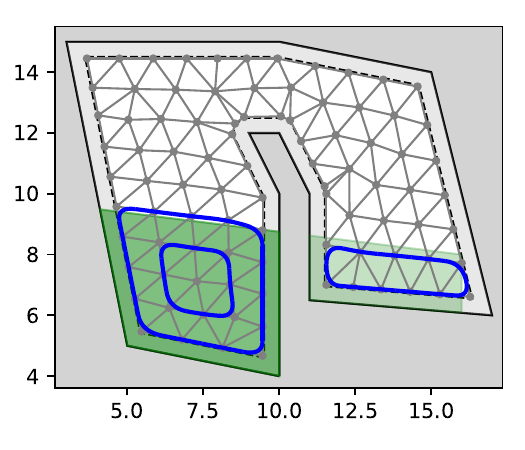}  
    \includegraphics[width=0.49\columnwidth]{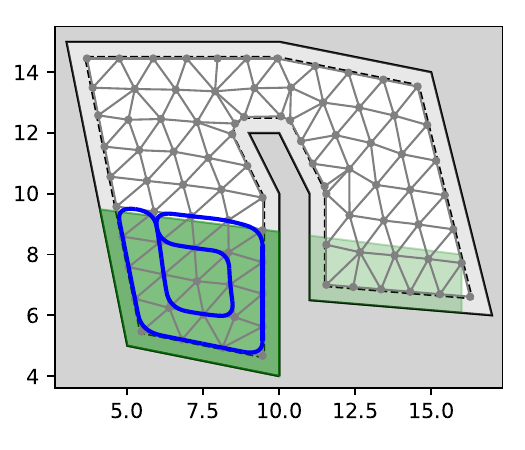}  
    \caption{Low value.}
  \end{subfigure}
  \begin{subfigure}[b]{0.49\columnwidth}
    \includegraphics[width=0.49\columnwidth]{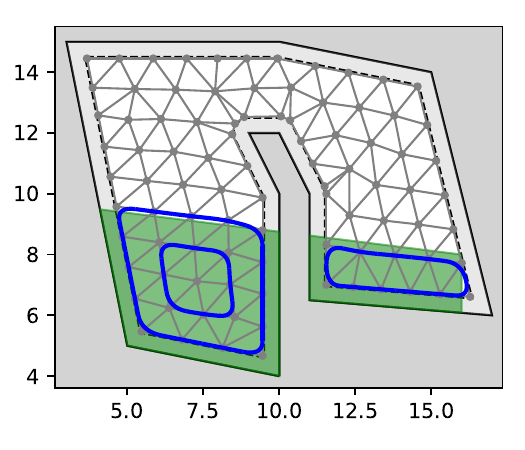}  
    \includegraphics[width=0.49\columnwidth]{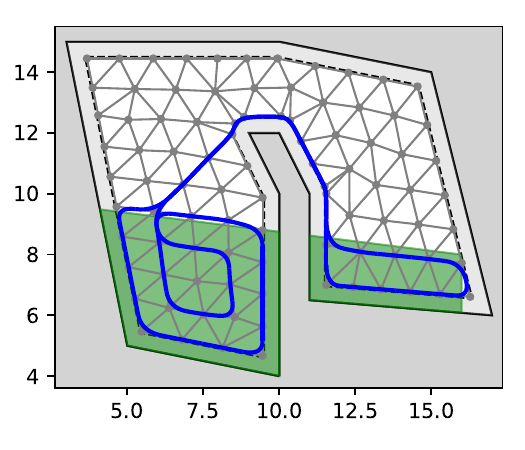}  
    \caption{Value increased.}
  \end{subfigure}
  \caption{If the valuable areas (green) are more distanced, the cycles (blue) should only be connected if the value is high enough in relation to the costs.
  In (a) the right area's value is not high enough and its cycle gets removed.
  In (b) the value is increased and the cycle gets connected.
  }\label{fig:cpp:prac:alg:connectpcst}
\end{figure}

Now, we only need to connect the cycles to form a tour.
For adjacent cycles, this is quite simple and involves only minimal extra costs: simply go through every edge that connects two cycles and perform a merge via the least expensive one, see \cref{fig:cpp:prac:alg:connect1}.
A simple optimization would be to use two parallel edges once, instead of one edge twice, but this is also done automatically in \cref{sec:cpp:prac:alg:topt}.
Things get more complicated if the cycles are farther apart.
It could be that the connection costs actually outweigh the touring costs of the corresponding cycles.
If the area covered by the cycle is not valuable enough, we are better off simply removing the cycle, see \cref{fig:cpp:prac:alg:connectpcst}.

To select any cycles, we first need to know how much each cycle is worth.
We estimate the value of a cycle by the sum of values of its covered waypoints.
If a waypoint occurs in different cycles, only the first cycle gets its value.
This can happen if two cycles cross and cannot be connected due to turn costs.
Because this rarely happens, the estimated cycle values are accurate if the values of the waypoints are accurate.
Otherwise, the value of a cycle can be underestimated and result in a slightly lower solution quality.

\begin{figure}[htb]
  \centering
  \includegraphics[width=0.9\columnwidth]{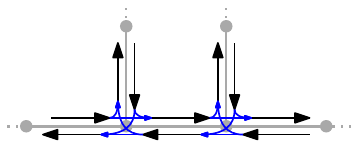}
  \caption{Converting the grid (gray) into a directed edge graph to compute a shortest path with turn costs inside.
  The distance and turn costs are assigned to the blue arcs.}\label{fig:cpp:prac:edgegraph}
\end{figure}

Next, we need to know how expensive it is to connect any two cycles.
This can be achieved with a Dijkstra-variant on the edge graph.
Working on the edge graph of the grid allows us to include not only the distance of the path, but also the turn costs between any two edges.
To make things simpler, we use a directed version where we also include the direction through which we pass the edge, as can be seen in \cref{fig:cpp:prac:edgegraph}.

The distance cost of using an edge can now simply be assigned to the outgoing arc in the edge graph.
If we let $k$ be the maximum degree in the grid, then we have at most $O(|P|\cdot k)$ vertices and $O(|P|\cdot k^2)$ edges in the auxiliary graph.
Using Dijkstra's algorithm, we can compute the least expensive path between any two edges (ignoring possibly collected coverage value) in $O(|P|\cdot k^2 \log |P|)$.
The costs are symmetric, so it is optimal in both directions.
Still missing are the costs involved merging a (doubled) path with a cycle.
It would be expensive to check all combinations for edges incident to the two cycles.
Instead, we can select one of the cycles and initialize all incident edges in the Dijkstra-algorithm with the final connection costs to it.
We now only have to find the least expensive incident edge to the target circle using the already computed distances by Dijkstra's algorithm.

With these two pieces of information, we can compute a prize-collecting steiner tree (PCST) on the cycles and their connections.
The resulting tree corresponds to the worthwhile cycles and how to connect them.
Computing an optimal PCST is \NP-hard, but the cycle covers obtained here are usually small enough to be solved optimally using integer programming.
Otherwise, an implementation~\cite{hegde2015nearly}  of  the  2-approximation by Goemans and Williamson~\cite{goemans1995general} can be used.
If there are some zero or negative connection costs, we can directly connect the corresponding cycles before we compute the PCST\@.
Using a PCST instead of just greedily connecting cycles potentially also integrates cycles that are not valuable enough on their own, but they become valuable in combination with other cycles.

Using the PCST, we now iteratively merge cycles (using the doubled paths computed using the Dijkstra-approach) in a depth-first search starting from an arbitrary cycle in the PCST\@.
Whenever we merge two cycles, the path creates additional docking points that may be cheaper than the originally computed connecting paths.
However, we do not need to recompute the whole Dijkstra-tree, but can simply reduce the costs for the corresponding edges and let the reduced costs propagate.
Caveat: During the joining of the cycle with the doubled path, passages are actually replaced from the cycle. 
The shortest paths originating from such removed passages become invalid.
As this rarely occurs and can be detected, recomputation should only be performed if such an invalid shortest path is about to be used.


\subsection[Local Optimization]{Step 7: Local Optimization}\label{sec:cpp:prac:alg:topt}

\begin{figure}[htb]
  \begin{subfigure}[b]{0.49\columnwidth}
    \includegraphics[width=\columnwidth]{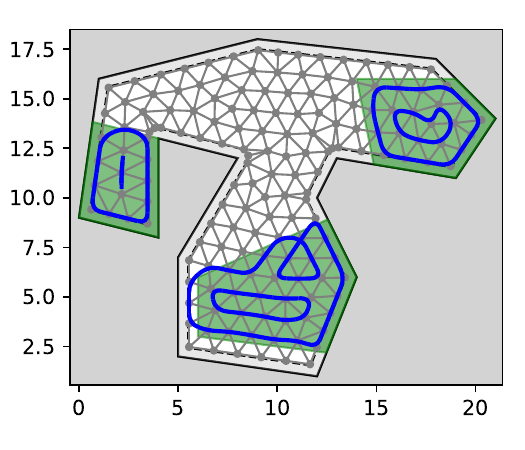}
    \caption{Cycles before connecting.}
  \end{subfigure}
  \begin{subfigure}[b]{0.49\columnwidth}
    \includegraphics[width=\columnwidth]{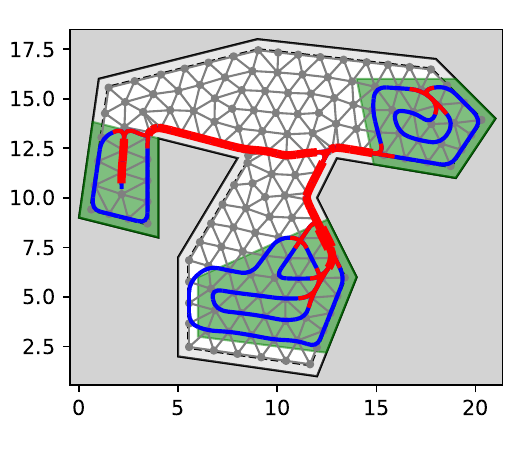}
    \caption{Tour.}
  \end{subfigure}
  \caption{Especially due to the connection approach of subtours, a lot of redundant coverages (red) can be created, which we aim to minimize.}\label{fig:cpp:prac:alg:touropt:redundant}
\end{figure}

After connecting the cycles to form a tour, the connecting parts are often highly redundant, as can be seen in \cref{fig:cpp:prac:alg:touropt:redundant}.
Luckily, we can extend the local optimization approach of \cref{sec:cpp:prac:alg:ccopt} to connected tours.
The challenge is to make sure that the tour remains connected after local optimizations.
The used MIP does not enforce connectivity and may disconnect the tour again.
A na\"ive approach is to only accept local improvements that preserve the connectivity and discard all others.
This is of course quite restrictive and we can find a superior solution.

Subtour elimination in the MIP is more difficult than for, e.g., the \textsc{Traveling Salesman Problem}: not only are all visitations optional, but two tours can cross without being connected.
Simply enforcing that two edges have to leave a connected component, therefore, does not yield the desired result.
In~\cite{icra_mosquito_2018} we actually have a corresponding MIP\@.
Because we already start with a tour and know that we have to connect an interior solution (inside the small area to be optimized) to the fixed exterior solution, we can devise a simpler separation constraint.

There are two types of subtours: those that are completely within the area and those that are only partially within the area.
We can only get an infeasible solution with subtours of the second type if the local solution incorrectly connects the exterior solution.
However, both types can be handled equally.

We either want a subtour $C$ to dissolve or to become part of the connected tour.
For this, there needs to be a vertex passage of a subtour to be unused, or a vertex passage leaving the subtour used.
We select an arbitrary vertex passage of the first type and demand that the sum of the second type is greater than it.
Note that this assumes the existence of an external, fixed solution, and is otherwise not exact.

Let $X_A$ be the vertex passage variables that are contained in the area $A$ and can be modified by the local optimization.
This includes all variable $x_{uvw}$ with $u,v,w\in A$.
If $u$ or $w$ are not in $A$, $vu$ resp. $vw$ must be used in the solution, i.e., the edge connects the changeable interior solution to the fixed exterior solution.
All other variables are fixed.

Let $X_C$ be the vertex passage variables that are used by the subtour $C$.
Let $X'_C$ be the vertex passage variables that share one edge with the subtour $C$ but are not in $X_C$.
These are the vertex passages that leave the trajectory of $C$.
We can now state a constraint that eliminates $C$, if it has been created by an optimization on $X_C$.

\begin{equation}
  \sum_{x\in X'_C\cap X_A} x \geq x_c \quad x_c \in X_C\cap X_A, \text{$C$ is subtour}
\end{equation}

\begin{figure}[htb]
  \includegraphics[width=\columnwidth]{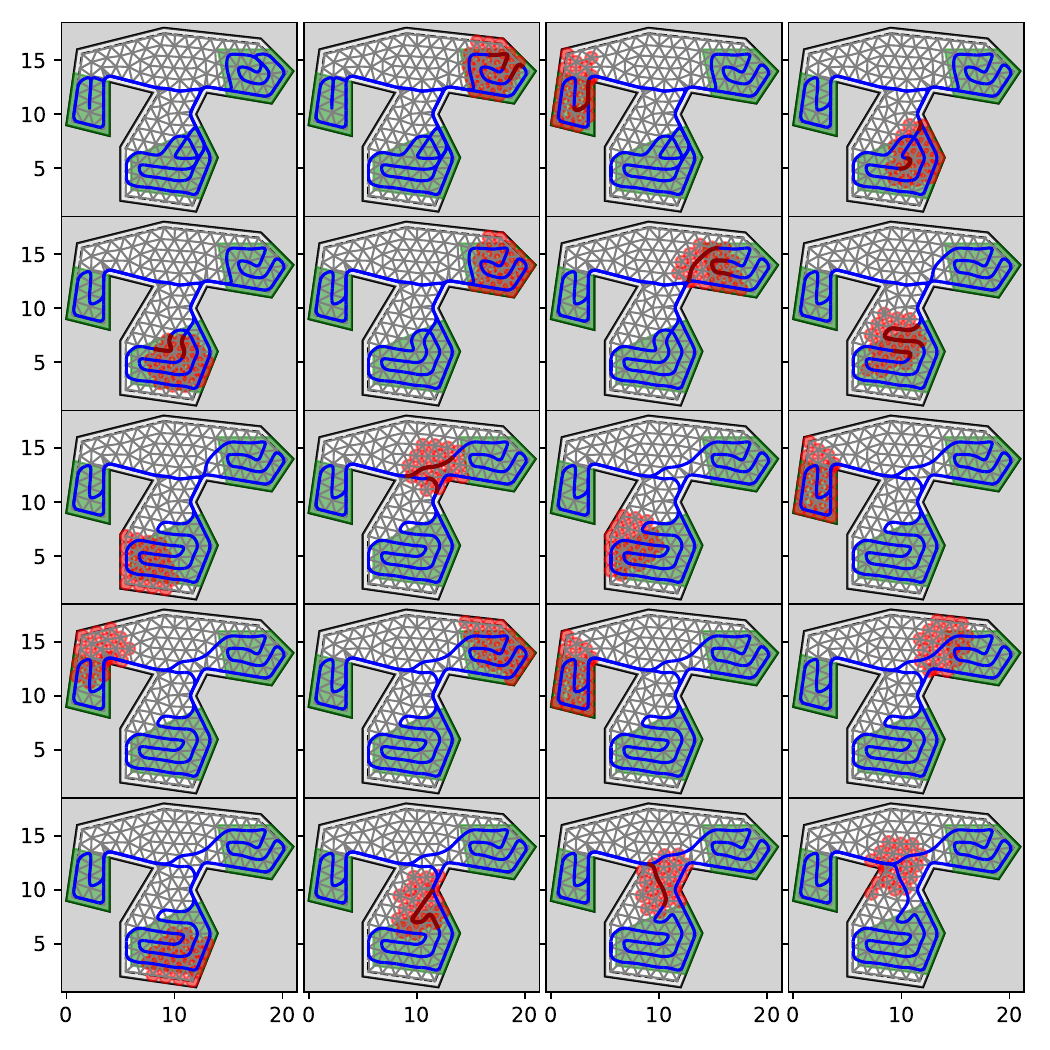}
  \caption{Multiple steps of the tour optimization.
  The optimized area and the changed parts are highlighted in red.
  In some steps, no changes are made because the solution is (locally) optimal in the area.}\label{fig:cpp:prac:alg:touropt:steps}
\end{figure}

There exist more efficient options for connecting, e.g., more distant subtours, but this hardly applies for optimizing only small areas.
For the case that the MIP does not yield a connected solution for $A$ within a fixed number of iterations, we discard the infeasible solution do not change $A$ in this iteration.
Applying this approach multiple times can significantly improve the solution, as can be seen in the example in \cref{fig:cpp:prac:alg:touropt:steps}.

\section{Evaluation}\label{sec:cpp:prac:opts}

In this section, we evaluate the performance of the algorithm on a set of benchmark instances.
We first evaluate the influence of the new optimizations and then the overall performance on the benchmark instances.

We generated \num{500} random instances for our benchmark using unions and differences of reasonably simple and thick random polygons.
The generation was supervised and the parameters were manually adjusted to create instances that mimic complex agricultural fields, architecture, groups of buildings, and other real-world scenarios.
The valuable areas and areas with increased costs were also chosen by randomly placing thick polygons, with possible overlaps that summed up.
Examples of these instance can be seen in \cref{fig:cpp:prac:grid:partialexamples}.
The selection of these examples was random and, thus, should reflect the distribution in the \num{500} instances.

All experiments were run on Ubuntu workstations with AMD Ryzen 7 5800X ($8\times\SI{3.8}{\GHz}$) CPU and \SI{128}{\giga\byte} of RAM\@.
The code is run with Python 3.8.8 and uses Gurobi 9.1.2.
More details can be found in the repository \url{https://github.com/d-krupke/ALENEX24-partial-coverage-path-planning}.

\subsection{Local Optimization}\label{sec:pcpp:exp:locopt}

\begin{figure}[tb]
  \begin{subfigure}[b]{0.49\columnwidth}
    \centering
    \includegraphics[width=0.9\columnwidth]{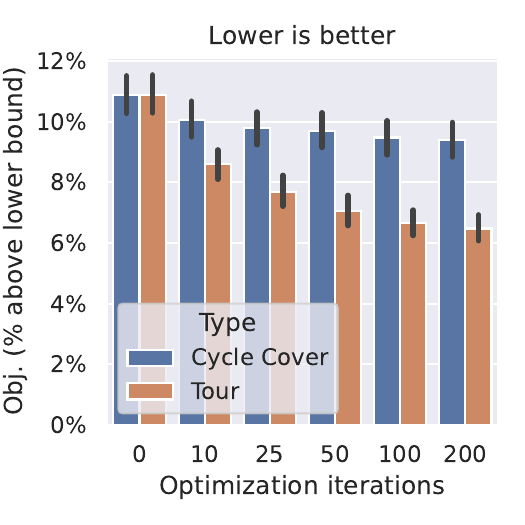}
    \caption{Iterations.}\label{fig:cpp:prac:eval:optstepspartial}
  \end{subfigure}
  \begin{subfigure}[b]{0.49\columnwidth}
    \centering
    \includegraphics[width=0.9\columnwidth]{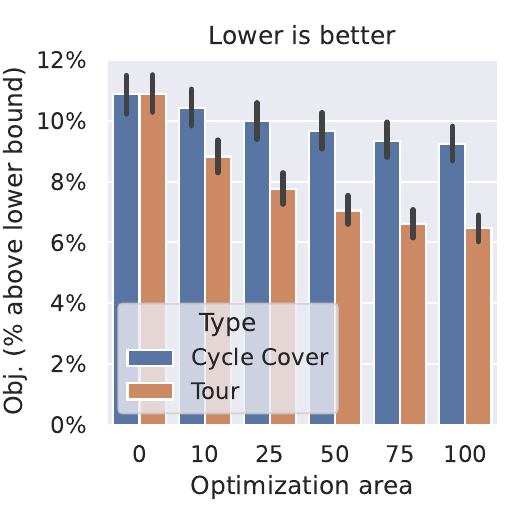}
    \caption{Optimization area.}\label{fig:cpp:prac:eval:optsareapartial}
  \end{subfigure}
  \caption{
    Influence of the number of iterations (a) and the optimization area (b) for the local optimization depending on whether it is applied on cycle covers (Step 5) or tours (Step 7).
    The $0$-bars are the baseline without local optimization and show the initial gap to the fractional solution (upper bound on optimality gap), the other bars show the optimality gap after the local optimization.
    This is the average over all \num{500} instances, the error bars show the standard deviation.
     } 
\end{figure}


In the first experiment, we evaluate the local optimization steps, that were not considered in the original algorithm.
The important questions are:
(1) How much can we improve the solution using this optimization?
We have to make sure that the improvement is worth the additional complexity.
(2) Should we focus on optimizing the cycle covers or the tours?
 While the tours are the final result, the cycle covers are less expensive to optimize.
(3) How much influence do the number of iterations and the size of the area have? 
The runtime increases linearly with the number of iterations but exponentially with the size of the area. 
However, the \NP-hard nature of the problem also implies that area cannot be fully substituted by iterations.

To answer these questions, we computed solutions that performed a local optimization on either cycle cover or tour with \numlist{0;10;25;50;100;200} iterations and an area of \num{50}~vertices.
Additionally, we computed solutions that performed \num{50} iterations of the local optimization on either cycle cover or tour, but with a varying area of \numlist{0;10;25;50;75;100} vertices.

The results in \cref{fig:cpp:prac:eval:optstepspartial} show that the optimizations with an area size of \num{50}~vertices yield a visible improvement for partial coverage in both steps.
\num{10}~iterations on the cycle cover already reduce the optimality gap (in comparison to the lower bound) by around \SI{10}{\percent}.
The further iterations lose effectiveness, as can be expected because we prioritize the expensive areas, but an improvement remains visible.
While the optimization is successful on cycle covers, it is even more impressive on tours.
Here, the first \num{10}~iterations lower the optimality gap by more than \SI{20}{\percent}.
The further iterations also remain stronger than for cycle cover, but their improvements still decline quickly.
This implies that the cycle covers are already nearly optimal, but the connection of the cycles to a tour is not very efficient.
The local optimization on tours can easily find (locally) suboptimal parts in the connected solution and improve them visibly.

The results in \cref{fig:cpp:prac:eval:optsareapartial} for varying area are surprisingly very similar: Doubling the area has a similar effect as quadrupling the iterations.
One difference is that for optimizing cycle covers, the larger areas are more important than for tours.
Optimizing only small areas with \num{10} vertices barely improves the solution.
For tours, on the other hand, such small areas can already make a significant difference.
This is very useful to know because optimizing \num{10}~vertices is extremely fast and could still be done by brute-force.
Thus, we can do many iterations with such small areas in a short time.
Larger areas still have their advantage, and \num{50}~iterations of size \num{100} are roughly as effective as \num{200}~iterations of size \num{50}.

\begin{figure}[tb]
  \begin{subfigure}[b]{0.49\columnwidth}
    \includegraphics[width=\columnwidth]{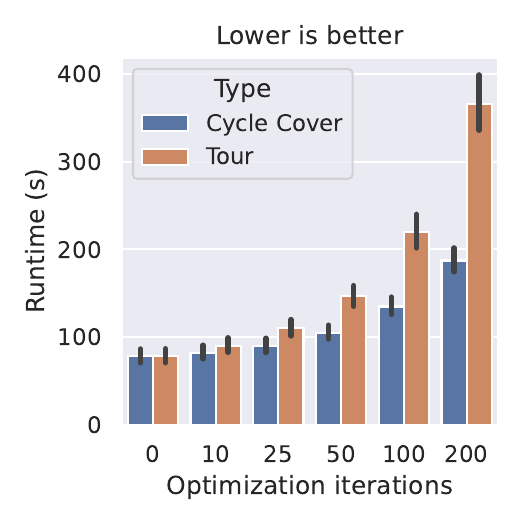}
    \caption{Iterations.}
  \end{subfigure}
  \begin{subfigure}[b]{0.49\columnwidth}
    \includegraphics[width=\columnwidth]{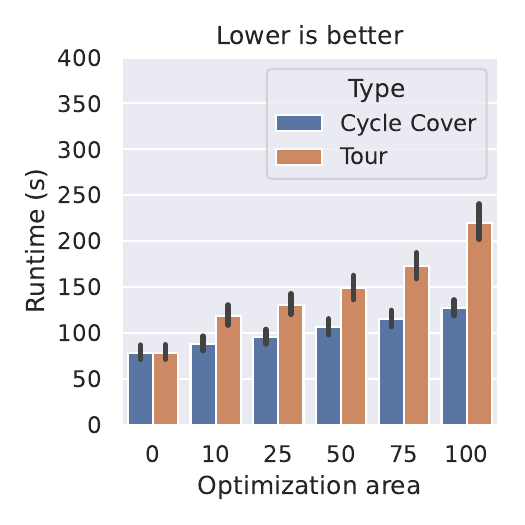}
    \caption{Optimization area.}
  \end{subfigure}
  \caption{Runtime for more iterations or larger areas in the local optimization.
  Shown as the mean runtime in seconds over all \num{500} instances.
   }\label{fig:cpp:prac:eval:optsruntime}
\end{figure}

The runtime differences for the number of iterations and the size of the area can be seen in \cref{fig:cpp:prac:eval:optsruntime}.
Surprisingly, the runtime for larger areas looks nearly linear (caveat: the x-axis for iterations is exponential, but it is almost linear for area).
However, this data should be used with caution because it can be skewed.
The implementation is only optimized for quality but not for runtime.
The connectivity detection, necessary to make sure that we did not accidentally disconnect the tour and have to insert constraints, is especially inefficient.
Instead of only analyzing the changed part, it always checks the whole solution with a procedure written in pure Python. 
This gives the tour variant a significant overhead which could be eliminated.
The tour variant will still remain slower because the solution frequently gets disconnected and needs to be reconnected using additional constraints.
These constraints become less efficient for larger areas because the solution develops more options to evade it.
For larger optimization areas, additional constraints should be developed and  used.

For the next experiment, we use \num{25} iterations of size \num{50} for both steps.
For tours, we use at most \num{10} cutting plane iterations.

\subsection{Optimality Gap}

\begin{figure}
  \begin{subfigure}[b]{0.49\columnwidth}
    \includegraphics[width=\columnwidth]{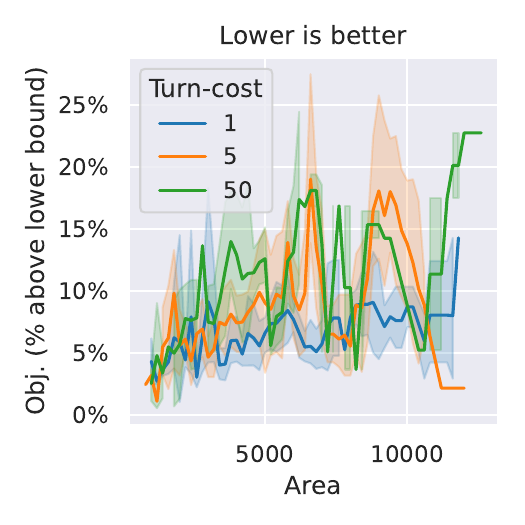}
    \caption{}\label{fig:cpp:prac:eval:optgaparea}  
  \end{subfigure}
  \begin{subfigure}[b]{0.49\columnwidth}
    \centering
    \includegraphics[width=\columnwidth]{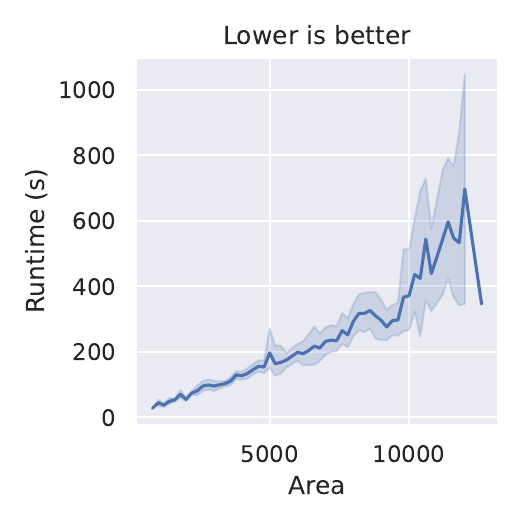}
    \caption{}\label{fig:cpp:prac:eval:runtime}
  \end{subfigure}
  \caption{(a) Optimality gap (compared to fractional solution as lower bound) on graph instance over area size.
  It shows that the optimality gap increases for larger instances, and that lower weights on turn costs result in better solutions.
  (b) Mean runtime over instance size of the Python-implementation. Note that it goes down again at the end because there are only few very large instances which by chance happen to be simpler than the average instance. Thus, this is a statistical effect and not a real trend.}
\end{figure}

To evaluate the overall performance  of the algorithm, we again run the algorithm on the \num{500} instances.
The plot in \cref{fig:cpp:prac:eval:optgaparea} shows how the quality of the solution develops over the size of the instances.
The quality is again measured by the difference of the objective value to the lower bound provided by the fractional solution (see \cref{sec:cpp:prac:alg:frac}).
We can see that the objective is around \SIrange{10}{15}{\percent} above the fractional solution, as we have already seen in the previous experiments.
However, we make the new observation here that the quality slightly degrades for larger instances.
Based on the tool radius of \num{1.0}, the larger (graph) instances have multiple thousand vertices.
This degradation could be converging, but the data is relatively noisy and has too small a range to make any sure assumptions.
The gap is generally smaller for lower turn costs, but this is not surprising because the turn costs make the problem combinatorially more complex.
This influences at least the quality of the fractional solution, which provides us with the lower bound.
Whether the actual solution has a larger optimality gap cannot be answered from the solution.

\subsection{Runtime}

The  primary focus of this  paper is on the quality of the solutions, but the runtime is also an important factor.
The original algorithm was able to solve instances with over \num{300000} vertices, though this could take several hours and require a powerful workstation.
The instances considered in this paper only have a few thousand vertices, as the implementation is only optimized for quality and not for runtime.
Despite being relatively small, the instances are still non-trivial, as seen in \cref{fig:cpp:prac:grid:partialexamples}.
These instances require a runtime of a few minutes, as can be seen in \cref{fig:cpp:prac:eval:runtime}.
Improving the efficiency of the prototype is possible in multiple places.
However, there are inherent challenges when compared to the original algorithm.
First, the original algorithm benefits from the simplicity of square grids, which have only three types of passages.
Second, it utilizes basic integer arithmetic, while the algorithm in this paper requires floating-point arithmetic, potentially affecting convergence behavior.
\section{Conclusion}\label{sec:pcpp:conclusion}

In this paper, we showed how to adapt a constant-factor approximation algorithm for coverage tours on grid graphs to arbitrary meshes derived from intricate, polygonal environments.
While the approximation factor may be lost in the process (if the mesh does not happen to be a perfect square grid), we demonstrated that the algorithm still yields low optimality gaps in practice.
Furthermore, we showcased its versatility in handling partial coverage and accommodating heterogeneous passage costs, offering the flexibility to trade off coverage quality and time efficiency.
This adaptation paves the way to compute efficient coverage paths with a robust theoretical foundation for real-world robotic applications.

Potential future work includes multi-robot variants of the problem, in which a fixed number of robots may be used.
The current approach should be extendable by only adapting the connection step (Step 6) if only the overall sum of costs is of interest.
If the individual costs are of interest, the proposed approach could be generalized by not only deciding for an orientation (Step 3) based on the linear relaxation,
but extending the linear relaxation to multiple robots (essentially copying it for every robot), and additionally deciding which robot should be used.
A practically relevant but algorithmically challenging variant is to maximize the coverage quality for a given budget.
Among others, one problem in our approach is the reliance on the fractional relaxation, which is known to be weak for budget constraints.
However, the linear relaxation could potentially be improved by additional constraints or by performing some branching steps.

\section*{Acknowledgments}
This work has been supported by the German Research Foundation
(DFG), project ``Computational Geometry: Solving Hard Optimization Problems'' (CG:SHOP),
grant FE407/21-1.


\bibliographystyle{plainurl}
\bibliography{bibliography}

\begin{thebibliography}{10}

\bibitem{aggarwal2000angular}
Alok Aggarwal, Don Coppersmith, Sanjeev Khanna, Rajeev Motwani, and Baruch
  Schieber.
\newblock The angular-metric traveling salesman problem.
\newblock {\em {SIAM} J. Comput.}, 29(3):697--711, 1999.
\newblock \href {https://doi.org/10.1137/S0097539796312721}
  {\path{doi:10.1137/S0097539796312721}}.

\bibitem{oswin2017minimization}
Oswin Aichholzer, Anja Fischer, Frank Fischer, J.~Fabian Meier, Ulrich
  Pferschy, Alexander Pilz, and Rostislav Stan{\v{e}}k.
\newblock Minimization and maximization versions of the quadratic travelling
  salesman problem.
\newblock {\em Optimization}, 66(4):521--546, 2017.

\bibitem{applegateconcorde}
David Applegate, Robert Bixby, Vasek Chvatal, and William Cook.
\newblock Concorde tsp solver.
\newblock URL: \url{http://www.math.uwaterloo.ca/tsp/concorde.html}.

\bibitem{applegate2011traveling}
David~L. Applegate, Robert~E. Bixby, Va{\v{s}}ek Chv{\'a}tal, and William~J.
  Cook.
\newblock {\em The traveling salesman problem}.
\newblock Princeton university press, 2011.

\bibitem{applegate2009certification}
David~L. Applegate, Robert~E. Bixby, Vasek Chv{\'{a}}tal, William~J. Cook,
  Daniel~G. Espinoza, Marcos Goycoolea, and Keld Helsgaun.
\newblock Certification of an optimal {TSP} tour through 85, 900 cities.
\newblock {\em Oper. Res. Lett.}, 37(1):11--15, 2009.
\newblock \href {https://doi.org/10.1016/j.orl.2008.09.006}
  {\path{doi:10.1016/j.orl.2008.09.006}}.

\bibitem{arkin2001optimal}
Esther~M. Arkin, Michael~A. Bender, Erik~D. Demaine, S{\'{a}}ndor~P. Fekete,
  Joseph S.~B. Mitchell, and Saurabh Sethia.
\newblock Optimal covering tours with turn costs.
\newblock In S.~Rao Kosaraju, editor, {\em Proceedings of the Twelfth Annual
  Symposium on Discrete Algorithms, January 7-9, 2001, Washington, DC, {USA}},
  pages 138--147. {ACM/SIAM}, 2001.
\newblock URL: \url{http://dl.acm.org/citation.cfm?id=365411.365430}.

\bibitem{arkin2005optimal}
Esther~M. Arkin, Michael~A. Bender, Erik~D. Demaine, S{\'{a}}ndor~P. Fekete,
  Joseph S.~B. Mitchell, and Saurabh Sethia.
\newblock Optimal covering tours with turn costs.
\newblock {\em {SIAM} J. Comput.}, 35(3):531--566, 2005.
\newblock \href {https://doi.org/10.1137/S0097539703434267}
  {\path{doi:10.1137/S0097539703434267}}.

\bibitem{ausiello2007prize}
Giorgio Ausiello, Vincenzo Bonifaci, Stefano Leonardi, and Alberto
  Marchetti-Spaccamela.
\newblock Prize collecting traveling salesman and related problems.
\newblock In Teofilo~F. Gonzalez, editor, {\em Handbook of Approximation
  Algorithms and Metaheuristics, Second Edition, Volume 1: Methologies and
  Traditional Applications}, pages 611--628. Chapman and Hall/CRC, 2018.
\newblock \href {https://doi.org/10.1201/9781351236423-34}
  {\path{doi:10.1201/9781351236423-34}}.

\bibitem{drone_vid}
Aaron~T. Becker, Mustapha Debboun, S{\'{a}}ndor~P. Fekete, Dominik Krupke, and
  An~Nguyen.
\newblock Zapping zika with a mosquito-managing drone: Computing optimal flight
  patterns with minimum turn cost (multimedia contribution).
\newblock In Boris Aronov and Matthew~J. Katz, editors, {\em 33rd International
  Symposium on Computational Geometry, SoCG 2017, July 4-7, 2017, Brisbane,
  Australia}, volume~77 of {\em LIPIcs}, pages 62:1--62:5. Schloss Dagstuhl -
  Leibniz-Zentrum f{\"{u}}r Informatik, 2017.
\newblock \href {https://doi.org/10.4230/LIPIcs.SoCG.2017.62}
  {\path{doi:10.4230/LIPIcs.SoCG.2017.62}}.

\bibitem{Bern04}
Marshall~W. Bern.
\newblock Triangulations and mesh generation.
\newblock In Jacob~E. Goodman and Joseph O'Rourke, editors, {\em Handbook of
  Discrete and Computational Geometry, Second Edition}, pages 563--582. Chapman
  and Hall/CRC, 2004.
\newblock \href {https://doi.org/10.1201/9781420035315.ch25}
  {\path{doi:10.1201/9781420035315.ch25}}.

\bibitem{BernP00}
Marshall~W. Bern and Paul~E. Plassmann.
\newblock Mesh generation.
\newblock In J\"org-R\"udiger Sack and Jorge Urrutia, editors, {\em Handbook of
  Computational Geometry}, pages 291--332. North Holland / Elsevier, 2000.
\newblock \href {https://doi.org/10.1016/b978-044482537-7/50007-3}
  {\path{doi:10.1016/b978-044482537-7/50007-3}}.

\bibitem{bormann2015new}
Richard Bormann, Joshua Hampp, and Martin H{\"{a}}gele.
\newblock New brooms sweep clean - an autonomous robotic cleaning assistant for
  professional office cleaning.
\newblock In {\em {IEEE} International Conference on Robotics and Automation,
  {ICRA} 2015, Seattle, WA, USA, 26-30 May, 2015}, pages 4470--4477. {IEEE},
  2015.
\newblock \href {https://doi.org/10.1109/ICRA.2015.7139818}
  {\path{doi:10.1109/ICRA.2015.7139818}}.

\bibitem{bormann2018indoor}
Richard Bormann, Florian Jordan, Joshua Hampp, and Martin H{\"{a}}gele.
\newblock Indoor coverage path planning: Survey, implementation, analysis.
\newblock In {\em 2018 {IEEE} International Conference on Robotics and
  Automation, {ICRA} 2018, Brisbane, Australia, May 21-25, 2018}, pages
  1718--1725. {IEEE}, 2018.
\newblock \href {https://doi.org/10.1109/ICRA.2018.8460566}
  {\path{doi:10.1109/ICRA.2018.8460566}}.

\bibitem{cabreira2019survey}
Tau{\~a}~M. Cabreira, Lisane~B. Brisolara, and Paulo~R. Ferreira~Jr.
\newblock Survey on coverage path planning with unmanned aerial vehicles.
\newblock {\em Drones}, 3(1), 2019.
\newblock URL: \url{https://www.mdpi.com/2504-446X/3/1/4}, \href
  {https://doi.org/10.3390/drones3010004} {\path{doi:10.3390/drones3010004}}.

\bibitem{cabreira2018energy}
Taua~M. Cabreira, Carmelo~Di Franco, Paulo Roberto~Ferreira Jr., and Giorgio~C.
  Buttazzo.
\newblock Energy-aware spiral coverage path planning for {UAV} photogrammetric
  applications.
\newblock {\em {IEEE} Robotics Autom. Lett.}, 3(4):3662--3668, 2018.
\newblock \href {https://doi.org/10.1109/LRA.2018.2854967}
  {\path{doi:10.1109/LRA.2018.2854967}}.

\bibitem{chen2011efficient}
Long Chen and Michael Holst.
\newblock Efficient mesh optimization schemes based on optimal delaunay
  triangulations.
\newblock {\em Computer Methods in Applied Mechanics and Engineering},
  200(9-12):967--984, 2011.

\bibitem{choset2001coverage}
Howie Choset.
\newblock Coverage for robotics - {A} survey of recent results.
\newblock {\em Ann. Math. Artif. Intell.}, 31(1-4):113--126, 2001.
\newblock \href {https://doi.org/10.1023/A:1016639210559}
  {\path{doi:10.1023/A:1016639210559}}.

\bibitem{choset2000exact}
Howie Choset, Ercan~U. Acar, Alfred~A. Rizzi, and Jonathan~E. Luntz.
\newblock Exact cellular decompositions in terms of critical points of morse
  functions.
\newblock In {\em Proceedings of the 2000 {IEEE} International Conference on
  Robotics and Automation, {ICRA} 2000, April 24-28, 2000, San Francisco, CA,
  {USA}}, pages 2270--2277. {IEEE}, 2000.
\newblock \href {https://doi.org/10.1109/ROBOT.2000.846365}
  {\path{doi:10.1109/ROBOT.2000.846365}}.

\bibitem{choset1998coverage}
Howie Choset and Philippe Pignon.
\newblock Coverage path planning: The boustrophedon cellular decomposition.
\newblock In {\em Field and service robotics}, pages 203--209. Springer, 1998.

\bibitem{de2011experimental}
Igor~R. de~Assis and Cid~C. de~Souza.
\newblock Experimental evaluation of algorithms for the orthogonal milling
  problem with turn costs.
\newblock In Panos~M. Pardalos and Steffen Rebennack, editors, {\em
  Experimental Algorithms - 10th International Symposium, {SEA} 2011,
  Kolimpari, Chania, Crete, Greece, May 5-7, 2011. Proceedings}, volume 6630 of
  {\em Lecture Notes in Computer Science}, pages 304--314. Springer, 2011.
\newblock \href {https://doi.org/10.1007/978-3-642-20662-7_26}
  {\path{doi:10.1007/978-3-642-20662-7_26}}.

\bibitem{douglas1973algorithms}
David~H. Douglas and Thomas~K. Peucker.
\newblock Algorithms for the reduction of the number of points required to
  represent a digitized line or its caricature.
\newblock {\em Cartographica: the international journal for geographic
  information and geovisualization}, 10(2):112--122, 1973.

\bibitem{du1999centroidal}
Qiang Du, Vance Faber, and Max~D. Gunzburger.
\newblock Centroidal voronoi tessellations: Applications and algorithms.
\newblock {\em {SIAM} Rev.}, 41(4):637--676, 1999.
\newblock \href {https://doi.org/10.1137/S0036144599352836}
  {\path{doi:10.1137/S0036144599352836}}.

\bibitem{ellefsen2017multiobjective}
Kai~Olav Ellefsen, Herman~Augusto Lepikson, and Jan~C. Albiez.
\newblock Multiobjective coverage path planning: Enabling automated inspection
  of complex, real-world structures.
\newblock {\em Appl. Soft Comput.}, 61:264--282, 2017.
\newblock \href {https://doi.org/10.1016/j.asoc.2017.07.051}
  {\path{doi:10.1016/j.asoc.2017.07.051}}.

\bibitem{CIAC2019}
S{\'{a}}ndor~P. Fekete and Dominik Krupke.
\newblock Covering tours and cycle covers with turn costs: Hardness and
  approximation.
\newblock In Pinar Heggernes, editor, {\em Algorithms and Complexity - 11th
  International Conference, {CIAC} 2019, Rome, Italy, May 27-29, 2019,
  Proceedings}, volume 11485 of {\em Lecture Notes in Computer Science}, pages
  224--236. Springer, 2019.
\newblock \href {https://doi.org/10.1007/978-3-030-17402-6_19}
  {\path{doi:10.1007/978-3-030-17402-6_19}}.

\bibitem{ALENEX19}
S{\'{a}}ndor~P. Fekete and Dominik Krupke.
\newblock Practical methods for computing large covering tours and cycle covers
  with turn cost.
\newblock In Stephen~G. Kobourov and Henning Meyerhenke, editors, {\em
  Proceedings of the Twenty-First Workshop on Algorithm Engineering and
  Experiments, {ALENEX} 2019, San Diego, CA, USA, January 7-8, 2019}, pages
  186--198. {SIAM}, 2019.
\newblock \href {https://doi.org/10.1137/1.9781611975499.15}
  {\path{doi:10.1137/1.9781611975499.15}}.

\bibitem{fischer2014exact}
Anja Fischer, Frank Fischer, Gerold J{\"{a}}ger, Jens Keilwagen, Paul Molitor,
  and Ivo Grosse.
\newblock Exact algorithms and heuristics for the quadratic traveling salesman
  problem with an application in bioinformatics.
\newblock {\em Discret. Appl. Math.}, 166:97--114, 2014.
\newblock \href {https://doi.org/10.1016/j.dam.2013.09.011}
  {\path{doi:10.1016/j.dam.2013.09.011}}.

\bibitem{galceran2013survey}
Enric Galceran and Marc Carreras.
\newblock A survey on coverage path planning for robotics.
\newblock {\em Robotics Auton. Syst.}, 61(12):1258--1276, 2013.
\newblock \href {https://doi.org/10.1016/j.robot.2013.09.004}
  {\path{doi:10.1016/j.robot.2013.09.004}}.

\bibitem{geuzaine2009gmsh}
Christophe Geuzaine and Jean-Fran{\c{c}}ois Remacle.
\newblock Gmsh: A 3-d finite element mesh generator with built-in pre-and
  post-processing facilities.
\newblock {\em International journal for numerical methods in engineering},
  79(11):1309--1331, 2009.

\bibitem{goemans1995general}
Michel~X. Goemans and David~P. Williamson.
\newblock A general approximation technique for constrained forest problems.
\newblock {\em {SIAM} J. Comput.}, 24(2):296--317, 1995.
\newblock \href {https://doi.org/10.1137/S0097539793242618}
  {\path{doi:10.1137/S0097539793242618}}.

\bibitem{hameed2014intelligent}
Ibrahim~A. Hameed.
\newblock Intelligent coverage path planning for agricultural robots and
  autonomous machines on three-dimensional terrain.
\newblock {\em J. Intell. Robotic Syst.}, 74(3-4):965--983, 2014.
\newblock \href {https://doi.org/10.1007/s10846-013-9834-6}
  {\path{doi:10.1007/s10846-013-9834-6}}.

\bibitem{hegde2015nearly}
Chinmay Hegde, Piotr Indyk, and Ludwig Schmidt.
\newblock A nearly-linear time framework for graph-structured sparsity.
\newblock In Francis~R. Bach and David~M. Blei, editors, {\em Proceedings of
  the 32nd International Conference on Machine Learning, {ICML} 2015, Lille,
  France, 6-11 July 2015}, volume~37 of {\em {JMLR} Workshop and Conference
  Proceedings}, pages 928--937. JMLR.org, 2015.
\newblock URL: \url{http://proceedings.mlr.press/v37/hegde15.html}.

\bibitem{itai1982hamilton}
Alon Itai, Christos~H. Papadimitriou, and Jayme~Luiz Szwarcfiter.
\newblock Hamilton paths in grid graphs.
\newblock {\em {SIAM} J. Comput.}, 11(4):676--686, 1982.
\newblock \href {https://doi.org/10.1137/0211056} {\path{doi:10.1137/0211056}}.

\bibitem{jager2008algorithms}
Gerold J{\"{a}}ger and Paul Molitor.
\newblock Algorithms and experimental study for the traveling salesman problem
  of second order.
\newblock In Boting Yang, Ding-Zhu Du, and Cao~An Wang, editors, {\em
  Combinatorial Optimization and Applications, Second International Conference,
  {COCOA} 2008, St. John's, NL, Canada, August 21-24, 2008. Proceedings},
  volume 5165 of {\em Lecture Notes in Computer Science}, pages 211--224.
  Springer, 2008.
\newblock \href {https://doi.org/10.1007/978-3-540-85097-7_20}
  {\path{doi:10.1007/978-3-540-85097-7_20}}.

\bibitem{jensen2020near}
Katharin~R. Jensen-Nau, Tucker Hermans, and Kam~K. Leang.
\newblock Near-optimal area-coverage path planning of energy-constrained aerial
  robots with application in autonomous environmental monitoring.
\newblock {\em {IEEE} Trans Autom. Sci. Eng.}, 18(3):1453--1468, 2021.
\newblock \href {https://doi.org/10.1109/TASE.2020.3016276}
  {\path{doi:10.1109/TASE.2020.3016276}}.

\bibitem{Kolmogorov2009}
Vladimir Kolmogorov.
\newblock Blossom {V:} a new implementation of a minimum cost perfect matching
  algorithm.
\newblock {\em Math. Program. Comput.}, 1(1):43--67, 2009.
\newblock \href {https://doi.org/10.1007/s12532-009-0002-8}
  {\path{doi:10.1007/s12532-009-0002-8}}.

\bibitem{lloyd1982least}
Stuart~P. Lloyd.
\newblock Least squares quantization in {PCM}.
\newblock {\em {IEEE} Trans. Inf. Theory}, 28(2):129--136, 1982.
\newblock \href {https://doi.org/10.1109/TIT.1982.1056489}
  {\path{doi:10.1109/TIT.1982.1056489}}.

\bibitem{marshall1994survey}
S.~Marshall and J.~G. Griffiths.
\newblock A survey of cutter path construction techniques for milling machines.
\newblock {\em The International Journal of Production Research},
  32(12):2861--2877, 1994.

\bibitem{mitchell1991weighted}
Joseph S.~B. Mitchell and Christos~H. Papadimitriou.
\newblock The weighted region problem: Finding shortest paths through a
  weighted planar subdivision.
\newblock {\em J. {ACM}}, 38(1):18--73, 1991.
\newblock \href {https://doi.org/10.1145/102782.102784}
  {\path{doi:10.1145/102782.102784}}.

\bibitem{modares2017ub}
Jalil Modares, Farshad Ghanei, Nicholas Mastronarde, and Karthik Dantu.
\newblock {UB-ANC} planner: Energy efficient coverage path planning with
  multiple drones.
\newblock In {\em 2017 {IEEE} International Conference on Robotics and
  Automation, {ICRA} 2017, Singapore, Singapore, May 29 - June 3, 2017}, pages
  6182--6189. {IEEE}, 2017.
\newblock \href {https://doi.org/10.1109/ICRA.2017.7989732}
  {\path{doi:10.1109/ICRA.2017.7989732}}.

\bibitem{murtaza2013priority}
Ghulam Murtaza, Salil~S. Kanhere, and Sanjay~K. Jha.
\newblock Priority-based coverage path planning for aerial wireless sensor
  networks.
\newblock In {\em 2013 {IEEE} Eighth International Conference on Intelligent
  Sensors, Sensor Networks and Information Processing, Melbourne, Australia,
  April 2-5, 2013}, pages 219--224. {IEEE}, 2013.
\newblock \href {https://doi.org/10.1109/ISSNIP.2013.6529792}
  {\path{doi:10.1109/ISSNIP.2013.6529792}}.

\bibitem{icra_mosquito_2018}
An~Nguyen, Dominik Krupke, Mary Burbage, Shriya Bhatnagar, S{\'{a}}ndor~P.
  Fekete, and Aaron~T. Becker.
\newblock Using a {UAV} for destructive surveys of mosquito population.
\newblock In {\em 2018 {IEEE} International Conference on Robotics and
  Automation, {ICRA} 2018, Brisbane, Australia, May 21-25, 2018}, pages
  7812--7819. {IEEE}, 2018.
\newblock \href {https://doi.org/10.1109/ICRA.2018.8463184}
  {\path{doi:10.1109/ICRA.2018.8463184}}.

\bibitem{oksanen2009coverage}
Timo Oksanen and Arto Visala.
\newblock Coverage path planning algorithms for agricultural field machines.
\newblock {\em J. Field Robotics}, 26(8):651--668, 2009.
\newblock \href {https://doi.org/10.1002/rob.20300}
  {\path{doi:10.1002/rob.20300}}.

\bibitem{papachristos2016distributed}
Christos Papachristos, Kostas Alexis, Luis Rodolfo~Garcia Carrillo, and Anthony
  Tzes.
\newblock Distributed infrastructure inspection path planning for aerial
  robotics subject to time constraints.
\newblock In {\em 2016 international conference on unmanned aircraft systems
  (ICUAS)}, pages 406--412. IEEE, 2016.

\bibitem{persson2004simple}
Per-Olof Persson and Gilbert Strang.
\newblock A simple mesh generator in {MATLAB}.
\newblock {\em {SIAM} Rev.}, 46(2):329--345, 2004.
\newblock \href {https://doi.org/10.1137/S0036144503429121}
  {\path{doi:10.1137/S0036144503429121}}.

\bibitem{rostami2013quadratic}
Borzou Rostami, Federico Malucelli, Pietro Belotti, and Stefano Gualandi.
\newblock Quadratic {TSP:} {A} lower bounding procedure and a column generation
  approach.
\newblock In Maria Ganzha, Leszek~A. Maciaszek, and Marcin Paprzycki, editors,
  {\em Proceedings of the 2013 Federated Conference on Computer Science and
  Information Systems, Krak{\'{o}}w, Poland, September 8-11, 2013}, pages
  377--384, 2013.
\newblock URL: \url{https://ieeexplore.ieee.org/document/6644028/}.

\bibitem{rowe2000finding}
Neil~C. Rowe and Robert~S. Alexander.
\newblock Finding optimal-path maps for path planning across weighted regions.
\newblock {\em Int. J. Robotics Res.}, 19(2):83--95, 2000.
\newblock \href {https://doi.org/10.1177/02783640022066761}
  {\path{doi:10.1177/02783640022066761}}.

\bibitem{nico_schlomer_2021_5196231}
Nico Schlömer, Antonio Cervone, G.~D. McBain, tryfon mw, Bin Wang, Nate, Filip
  Gokstorp, Ruben van Staden, toothstone, Jørgen~Schartum Dokken, Andrew
  Micallef, Dominic Kempf, Juan Sanchez, Keurfon Luu, anzil, Matthias
  Bussonnier, Raul~Ciria Aylagas, Fred Fu, ivanmultiwave, Nils Wagner, Siwei
  Chen, tayebzaidi, Tomislav Maric, Amine Aboufirass, and Yuan Feng.
\newblock nschloe/pygmsh: None, aug 2021.
\newblock \href {https://doi.org/10.5281/zenodo.5196231}
  {\path{doi:10.5281/zenodo.5196231}}.

\bibitem{nico_schlomer_2021_4728056}
Nico Schlömer and Adam Dobrawy.
\newblock nschloe/optimesh: None, apr 2021.
\newblock \href {https://doi.org/10.5281/zenodo.4728056}
  {\path{doi:10.5281/zenodo.4728056}}.

\bibitem{nico_schlomer_2021_5019221}
Nico Schlömer and J.~Hariharan.
\newblock nschloe/dmsh: None, jun 2021.
\newblock \href {https://doi.org/10.5281/zenodo.5019221}
  {\path{doi:10.5281/zenodo.5019221}}.

\bibitem{sharma2019optimal}
Gokarna Sharma, Ayan Dutta, and Jong-Hoon Kim.
\newblock Optimal online coverage path planning with energy constraints.
\newblock In Edith Elkind, Manuela Veloso, Noa Agmon, and Matthew~E. Taylor,
  editors, {\em Proceedings of the 18th International Conference on Autonomous
  Agents and MultiAgent Systems, {AAMAS} '19, Montreal, QC, Canada, May 13-17,
  2019}, pages 1189--1197. International Foundation for Autonomous Agents and
  Multiagent Systems, 2019.
\newblock URL: \url{http://dl.acm.org/citation.cfm?id=3331820}.

\bibitem{soltero2014decentralized}
Daniel~E. Soltero, Mac Schwager, and Daniela Rus.
\newblock Decentralized path planning for coverage tasks using gradient descent
  adaptive control.
\newblock {\em Int. J. Robotics Res.}, 33(3):401--425, 2014.
\newblock \href {https://doi.org/10.1177/0278364913497241}
  {\path{doi:10.1177/0278364913497241}}.

\bibitem{voros1999mobile}
Jozef V{\"{o}}r{\"{o}}s.
\newblock Mobile robot path planning among weighted regions using quadtree
  representations.
\newblock In Franz Pichler, Roberto Moreno-D{\'i}az, and Peter Kopacek,
  editors, {\em Computer Aided Systems Theory - EUROCAST'99, Vienna, Austria,
  September 29 - October 2, 1999, Proceedings}, volume 1798 of {\em Lecture
  Notes in Computer Science}, pages 239--249. Springer, 1999.
\newblock \href {https://doi.org/10.1007/10720123_20}
  {\path{doi:10.1007/10720123_20}}.

\bibitem{ware2016analysis}
John Ware and Nicholas Roy.
\newblock An analysis of wind field estimation and exploitation for quadrotor
  flight in the urban canopy layer.
\newblock In Danica Kragic, Antonio Bicchi, and Alessandro~De Luca, editors,
  {\em 2016 {IEEE} International Conference on Robotics and Automation, {ICRA}
  2016, Stockholm, Sweden, May 16-21, 2016}, pages 1507--1514. {IEEE}, 2016.
\newblock \href {https://doi.org/10.1109/ICRA.2016.7487287}
  {\path{doi:10.1109/ICRA.2016.7487287}}.

\bibitem{zheng2010multirobot}
Xiaoming Zheng, Sven Koenig, David Kempe, and Sonal Jain.
\newblock Multirobot forest coverage for weighted and unweighted terrain.
\newblock {\em {IEEE} Trans. Robotics}, 26(6):1018--1031, 2010.
\newblock \href {https://doi.org/10.1109/TRO.2010.2072271}
  {\path{doi:10.1109/TRO.2010.2072271}}.

\end{thebibliography}

\appendix{}

\section{Advanced Problem Definition}\label{sec:cpp:prac:problemdef}

The implementation actually has a more complex problem definition in mind as the page limit allows us to discuss.
In the following, we discuss the full underlying problem definition and the motivation behind it, as well as some more details on the discretization.
This section should not be required to understand the paper, but it sheds some more light on the intentions behind the implementation.

\subsection{Geometric Model}

We evaluate our optimization approach on a simplified, but still generic two-dimensional geometric model, which we define in this section.
This model can be adapted to many realistic scenarios, and many specifications are not due to algorithmic limitations but only used to simplify the evaluation.
While a simulation based evaluation would yield more realistic results, it would be less generic and require a large set of realistic instances, which is hard to come by.

Let us first discuss how we model the robot.
In the following, we primarily speak of robots, but generally all kinds of tools like milling machines or UAVs are included.
We model the robot as a circle of radius $r>0$, and its position $p\in \mathbb{R}^2$ is defined by its middle point.
The robot immediately covers everything below it, i.e., if it is at position $p$, $\textsc{Cov}(p)=\{p' \in \mathbb{R}^2 \mid ||p-p'||\leq r\}$ is covered.
This makes the robot rotational invariant and simplifies many computations.
The circular coverage may seem unrealistic at first glance for, e.g., a mower; but in a tour, the coverage of a line perpendicular to the trajectory is nearly identical to that of a circle.

The environment, e.g., walls or obstacles, can restrict the robot's movement.
We denote the \emph{feasible area} $F\subset \mathbb{R}^2$ as the set of all feasible positions of the robot and approximate it by a (non-simple) polygon.
In the examples and evaluations, we start with a larger polygon representing, e.g., a room, and shrink it by removing the parts too close to the boundaries.
$F$ does not need to coincide with the coverable area, which allows us to separate the robot's shape from its coverage.

We define the trajectory $T$ of the robot as a closed chain of waypoints $w_0, w_1, \ldots w_{|T|-1} \in F$.
The robot moves in straight lines between the waypoints.
We denote the corresponding segments by $\textsc{Segments}(T)=w_0w_1,w_1w_2,\ldots,w_{|T|-1}w_0$ and demand that all segments $s\in \textsc{Segments}(T)$ are fully contained in the feasible area $F$.
The trajectory in the following part is also called \emph{tour}.
An intermediate solution of multiple (closed) trajectories that still have to be connected to a tour is called a \emph{cycle cover}, and its elements are \emph{cycles} or \emph{subtours}.

Additional to the feasible area $F$, we have \emph{valuable areas} and \emph{expensive areas}.
Valuable areas $\mathcal{Q}=Q_0,Q_1,\ldots \subset \mathbb{R}^2$ with weights $t(Q_i)\in \mathbb{R}^+$ represent the parts we want to cover.
Expensive areas $\mathcal{E}=E_0, E_1, \ldots \subset F$ with weights $m(E_i)\in \mathbb{R}^+$ represent areas with increased touring costs.
Both types of areas are again approximated by polygons to simplify the computations.

The objective is to compute a feasible tour that maximizes the coverage value and minimizes the touring costs.
To combine maximizing the coverage and minimizing the costs, we convert the maximization of the coverage into a minimization.
This is achieved by considering the opportunity loss, i.e., the value of the missed area.
\footnote{We use $|Q|$ to denote the size (area) of a polygon $Q$ and the pure $Q$ to denote the set of all contained points.}
\begin{equation}
  \label{eq:pcpp:obj}
  \min_T \textsc{CoverageLoss}_{\mathcal{Q}, r}(T) + \textsc{TourCost}_\mathcal{E}(T)
\end{equation}
\[\text{s.t.} \quad s\subseteq F \quad \forall s\in\textsc{Segments}(T)\]
The advantage of this objective over others is that its lower bound is zero, which allows a better comparison.
We define the coverage loss and touring costs in the following.

Let $C_r(T)=\{p\in \mathbb{R}^2 | \exists s\in \textsc{Segments}(T), p' \in s: p \in \textsc{Cov}(p')\}$ denote the covered area of a tour.
Note that the inclusion of a point in a segment is defined as lying anywhere on the segment and not just on its endpoints.
This allows us to define the coverage loss formally by the maximally achievable coverage value minus the actually achieved value.
\begin{align*}
\textsc{CoverageLoss}_{\mathcal{Q}, r}(T) =& \sum_{Q\in \mathcal{Q}} |Q|\cdot t(Q) \\&- \sum_{Q\in \mathcal{Q}} |Q\cap C_r(T)|\cdot t(Q)
\end{align*}

The touring costs consist of weighted distances and turn angles.
\begin{align}\label{eq:pcpp:def:tourcost}
\textsc{TourCost}_\mathcal{E}(T)=& \lambda_0\cdot \textsc{DistCost}_\mathcal{E}(T)\\ \nonumber &+\lambda_1 \cdot \textsc{TurnCost}_\mathcal{E}(T)
\end{align}
The two weights $\lambda_0, \lambda_1 \geq 0$ allow weighting the distance and tour costs, and we vary them in our experiments.
Let $\mu_\mathcal{E}: \mathbb{R}^2\rightarrow \mathbb{R}^+$ define the cost multiplier at a tool position, which allows us to model local cost changes induced by the environment.
It is computed by $\mu_\mathcal{E}(p)=\prod_{E\in \mathcal{E}, p\in E} m(E)$.

The distance costs are now defined as
\begin{equation}
  \label{eq:pcpp:def:distcost}
\textsc{DistCost}_{\mathcal{E}}(T)= \sum_{s\in \textsc{Segments}(T)}\int_{p\in s}\mu_\mathcal{E}(p)\,dp\
\end{equation}
For $\mathcal{E}=\emptyset$, this becomes $\sum_{s\in \textsc{Segments}(T)}||s||$.
The turn costs only occur at waypoints and are also subject to the multiplier.
\begin{equation}
  \label{eq:pcpp:def:turncost}
\textsc{TurnCost}_\mathcal{E}(T) = \sum_{i=0}^{|T|-1} \mu_\mathcal{E}(w_i)\cdot\textsc{Turn}(w_{i-1}, w_i, w_{i+1})
\end{equation}
$\textsc{Turn}(p_0, p_1, p_2)$ denotes the turn angle at $p_1$ while traversing $p_0\rightarrow p_1 \rightarrow p_2$.
The indices of the waypoints are taken modulo $|T|$ to form a cycle.

\subsection[Discretization]{Discretization}

Before we can apply our approximation technique, we have to convert the polygonal area into a graph of potential waypoints.
Instead of a complex geometric problem, we then just have to find a tour in a graph where each vertex yields some coverage, and the touring costs are based on the used edges and edge transitions.

\begin{figure}[tbp]
  \begin{subfigure}[b]{0.3\columnwidth}
    \includegraphics[width=\columnwidth]{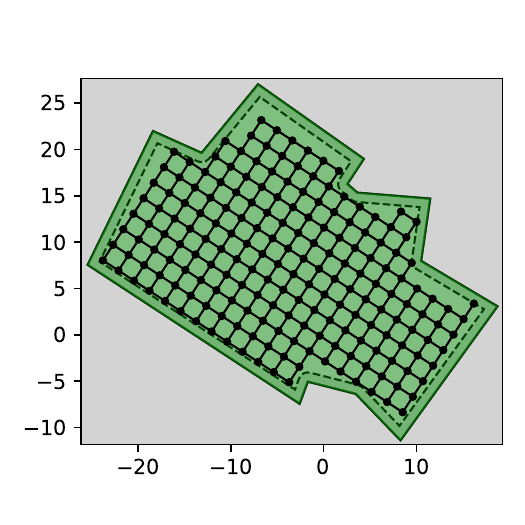}
    \caption{Square grid.}
  \end{subfigure}
  \hfill
  \begin{subfigure}[b]{0.3\columnwidth}
    \includegraphics[width=\columnwidth]{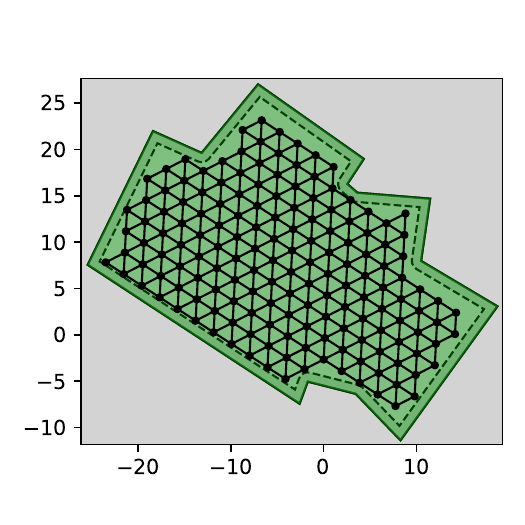}
    \caption{Triang. grid.}
  \end{subfigure}
  \hfill
  \begin{subfigure}[b]{0.3\columnwidth}
    \includegraphics[width=\columnwidth]{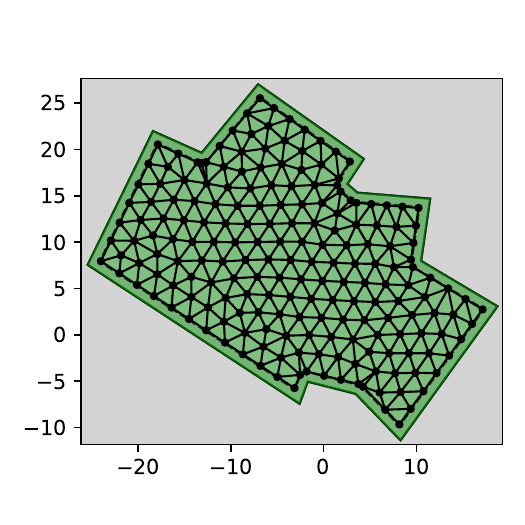}
    \caption{Mesh.}\label{fig:cpp:practical:grids:mesh}
  \end{subfigure}
  \caption[Different grids for transforming a polygon to a graph]{%
    Different grids for transforming a polygonal instance to a graph instance that can be solved with a variation of the approximation algorithm.
    The green area is the area to be covered but due to the robot's radius, we can only place waypoints inside the dashed area.
    We can rotate a regular grid to fit especially the interior area nicely.
    Alternatively, we can use an irregular grid created by a meshing algorithm.
    It can better adapt to the shape (especially the boundary) of the area, but its irregularity can also make it more expensive for coverage inside the polygon.
    }\label{fig:cpp:practical:grids}
\end{figure}

The simplest and most common strategy is to place a regular square grid over the feasible area.
The points and edges that are fully contained, become our graph.
This would also directly allow us to use the algorithm of Fekete and Krupke~\cite{ALENEX19}.
However, this is not always optimal.
Other options are to use regular triangular grids or irregular generated meshes.
To keep the computational costs low, it is generally better to have fewer vertices and a low edge degree at the vertices.
Not only the computational costs increase, also the quality of the relaxation decreases at vertices with more neighbors.
Examples with various grids can be seen in \cref{fig:cpp:practical:grids}.

\begin{figure}[tbp]
  \begin{subfigure}[b]{0.3\columnwidth}
    \includegraphics[width=\columnwidth]{./figures/experiments/02_algorithm_explanation/03a_instance.pdf}
    \caption{Instance with graph.}
  \end{subfigure}
  \hfill
  \begin{subfigure}[b]{0.3\columnwidth}
    \includegraphics[width=\columnwidth]{./figures/experiments/02_algorithm_explanation/03b_voronoi.pdf}
    \caption{Voronoi cells.}
  \end{subfigure}
  \hfill
  \begin{subfigure}[b]{0.3\columnwidth}
    \includegraphics[width=\columnwidth]{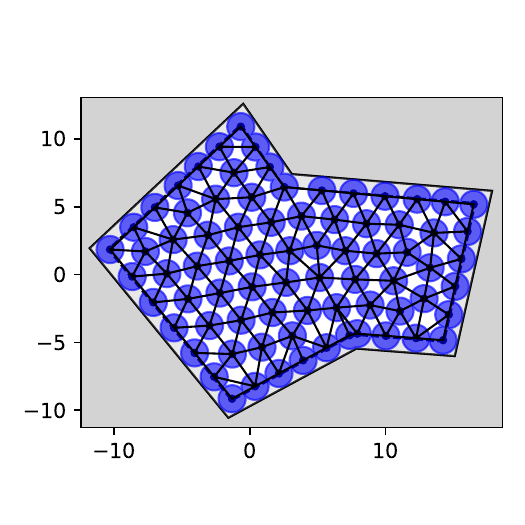}
    \caption{Robot's coverage at waypoints.}
  \end{subfigure}
  \caption[Computing the value of a vertex]{%
    Computing the value of a vertex, especially when using irregular grids, should use Voronoi diagrams or similar techniques to obtain a nice partition of the area, as seen in (b).
    Using only the area covered at the corresponding waypoint yields under- and overestimation due to ignored area that would indirectly be covered by using edges and intersecting ranges, as  seen in (c).
    The corresponding instance with the graph is shown in (a).
    Underestimating the value of a waypoint can result in the algorithm skipping it.
    Overestimating the value of a waypoint can result in the algorithm including it at a high cost.
    }\label{fig:cpp:practical:coveragevalueatpoint}
\end{figure}

Computing the edge costs and the turn costs at the vertices is straight-forward and can directly use the definition in \cref{eq:pcpp:def:tourcost}.
The graph is just a subset of the actual solution space, and so we can simply precompute the costs of the individual parts used by the graph.
The value of covering a vertex is more complicated.
We can simply assign the value that the robot covers when being at this point, but this can easily over- or underestimate the real value.
It can underestimate the value if the real coverage actually happens when moving to and from the vertex.
It can overestimate the value if other vertices are close by, and the coverage areas are overlapping.
Generally, it is good if the sum of values in the graph also equals the maximal value in the original instance.
We could simply scale all the values to achieve this but as the graph can have a heterogeneous distribution, using a Voronoi diagram is a better option.
The Voronoi diagram is a classical method from Computational Geometry which partitions the area such that each vertex gets the area assigned closest to it.
Using the value of these areas gives us a value assignment that equals the original values and is sensitive to the neighborhood of the vertices.
This can be seen in \cref{fig:cpp:practical:coveragevalueatpoint}.

In the following, we denote the resulting graph by $G=(P,E)$ and call $P$ (potential) waypoints.
Every tour $T$ on $G$ consists of segments in $E$, which are fully contained in the feasible area, and is, thus, a feasible tour in the original polygon.
We denote the coverage value assigned to a waypoint $p\in P$ by $\text{val}(p)$, and it corresponds to the coverage value of $p$'s Voronoi cell.
The distance cost of an edge $pp'\in E$ is defined by $\text{dist}(p,p')=\int_{x\in pp'}\mu_\mathcal{E}(x)\,dx$, according to \cref{eq:pcpp:def:distcost}.
The turn cost of passing $p$ through the neighbors $n$ and $n'\in N(p)$ is defined by $\text{turn}(n,p,n')=\mu_\mathcal{E}(p)\cdot\textsc{Turn}(n, p, n')$ according to \cref{eq:pcpp:def:turncost}.

Obtaining a good graph is a fundamental problem, and the whole \cref{sec:cpp:prac:grid} is focused on it.
\section{Implementation Details of Step 3}\label{apdx:cpp:prac:step3}

  In this step, we convert the instance such that we can use a minimum-weight perfect matching to compute an integral cycle cover, i.e., a solution, that is allowed to consist of multiple tours.
  Without turn costs, an optimal cycle cover can actually be computed in polynomial time because the costs of the edges are independent.
  With turn costs, the cost of an edge depends on orientation of the preceding edge, making the problem \NP-hard even in grid graphs~\cite{CIAC2019}.
  We use the fractional solution of the previous step to predict the corresponding orientation and make the costs independent again.
  
\begin{figure}
    \centering
    \includegraphics[width=0.7\columnwidth]{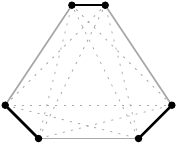}
    \caption{%
      Segments (black) can be connected to a cycle cover via a minimum-weight perfect matching (gray) on the end points.
      Because of the fixed joints, we can charge the turn costs to the edge weights.
      }\label{fig:cpp:prac:alg:segmentmatching}
  \end{figure}

  We can imagine this procedure as replacing every waypoint by an epsilon-length segment, as in \cref{fig:cpp:prac:alg:atomicstripsexample}.
  Computing a minimum-weight perfect matching on the endpoints, as in \cref{fig:cpp:prac:alg:segmentmatching}, yields the optimal cycle cover that includes all these segments.
  The necessary turns at the joints are fixed for every connecting edge, and can therefore be accounted for in the edge weights together with the distance.
  We are calling these epsilon-length segments \emph{atomic strips}.
  The possibility of skipping a waypoint can be implemented by adding an edge between the two endpoints of its atomic strip with the weight of the missed coverage.
  
  The orientations of the atomic strips are of fundamental importance:
  If we guess them correctly, the minimum-weight perfect matching actually corresponds to an optimal cycle cover on the waypoints.
  If we guess the orientation of an atomic strip badly, the minimum-weight perfect matching may perform expensive turns to integrate it.
  
\begin{figure}
    \centering
    \includegraphics[width=0.5\columnwidth]{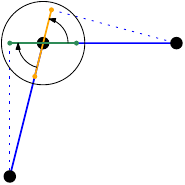}
    \caption{The orange and the green atomic strips represent the turn equally well.
    Only the assignment of the turn costs to the weight of the matching edges changes.
    Also, all atomic strips in between are equally good.}\label{fig:cpp:prac:alg:atomicstripsrotation}
  \end{figure}

  Luckily, the exact orientation is less important if we make turns at a waypoint, as the range of optimal orientations increases with the turn angle, as shown in \cref{fig:cpp:prac:alg:atomicstripsrotation}.
  For a U-turn, every orientation is optimal.
  The straighter a passage, the more important a good orientation becomes; but often these cases are easy to guess from the fractional solution. 
  
  This observation allows us to limit the orientations to the orientations of incident edges, i.e., neighbors.
  In the following, we represent the available orientations for the atomic strip of a waypoint $p\in P$ by the adjacent waypoints $N(p)$.
  If the atomic strip of $p$ has the orientation $n\in N(p)$, one of its endpoints heads at $n$.
  
\begin{figure}
    \centering
    \includegraphics[width=0.7\columnwidth]{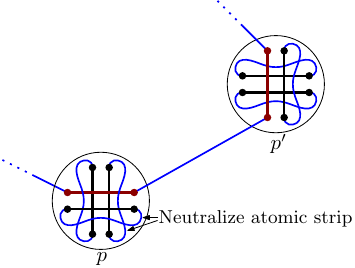}
    \caption{The waypoints $p$ and $p'$ each have two horizontal and two vertical atomic strips.
    The tour induced by a matching (blue) only uses one atomic strip of each, and skips the other by edges connecting both endpoints.
    These edges are usually zero-weight, except for one carefully-selected one that has the weight of the opportunity loss $\text{val}(p)$ resp. $\text{val}(p')$.
    }\label{fig:cpp:prac:alg:multipleatomicstrips}
  \end{figure}

  A waypoint may need to be crossed multiple times, as we are limited to passages within the grid $G=(P,E)$.
  This can easily be implemented by transitive edges, i.e., two successive edges $uv$ and  $vw\in E$ automatically create an edge $uw$ with the combined costs.
  However, we learned in \cite{ALENEX19} that introducing \emph{optional} atomic strips and only allowing direct connections scales much better.
  An optional atomic strip can be implemented by simply adding an edge with zero weight between its endpoints, see \cref{fig:cpp:prac:alg:multipleatomicstrips}.
  We call the non-optional atomic strip of a waypoint the \emph{dominant} one.
  
  In a square grid as in Fekete and Krupke~\cite{ALENEX19}, we obtain a 4-approximation if we add an atomic strip for every neighbor and declare the most used one in the fractional solution as dominant.
  In a triangular grid, there can be waypoints that are passed a linear number of times, as in \cref{fig:grid:hexaAtomicProblem}, but this is an artificial instance.
  In our instances, every waypoint is usually only covered once or twice.
  As every atomic strip increases the computational complexity, we limit the number of atomic strips to a constant $k$, and allow for every waypoint $p\in P$ at most one atomic strip per neighbor $n\in N(p)$.
  The task is to select a subset $A \subseteq N(p)$ with $|A|\leq k$ as atomic strips and determine the dominant one.
  
  If $k\geq |N(p)|$, we can simply choose $A=N(p)$.
  This allows us to use any passage twice without overhead, because any waypoint passage has either two neighbors with each having an optimal atomic strip or the passage is a U-turn.
  If $k<|N(v)|$, things get more complicated because we want to optimize three often opposing objectives:
  \begin{itemize}
    \item We want to improve the expected case, i.e., the passages with the highest likelihood should be as cheap as possible.
    \item We want to minimize the cost overhead of the average case, i.e., the average overhead of any passage.
    \item We want to minimize the worst case, i.e., the cost of the worst unexpected passage.
  \end{itemize}
  
  Our strategy for this case consists of two phases.
  First, we select atomic strips based on edge usages in the fractional solution.
  This optimizes the expected case.
  Second, we fill up the remaining atomic strips by minimizing the sum of squared overheads of passages not used in the fractional solution.
  This optimizes the average and worst case scenarios (using a higher exponent would shift the focus to the worst case).
  
  The precise strategy is given in Alg.~\ref{alg:pcpp:atomicstripselection};
  $\textsc{FS}(v,w)=\sum_{u\in N(v)} x_{uvw}$ denotes the usage of the edge $vw\in E$ in the fractional solution, and $\text{OH}(uvw, A)$ denotes the minimal overhead if the passage $uvw$ has to use an atomic strip in $A\subseteq N(v)$.
  The overhead corresponds to the additional turn costs needed to accommodate a (possibly misaligned) atomic strip.
  \begin{algorithm2e}
    \DontPrintSemicolon{}
    \KwInput{A waypoint $v$ and number $k$.}
    \KwOutput{A selection of at most $k$ atomic strips, identified by $N(v)$.}
    $A \gets \emptyset$\;
    $C_0 \gets \{n\in N(v) \mid \textsc{FS}(v,n)\geq \varepsilon\}$\;
    $C_1 \gets N(v)\setminus C_0$\;
  
    \tcc{1. Select by usage in fractional solution.}
    \While{$C_0\setminus A\not=\emptyset$}{
      $A \gets A+\text{argmax}_{v\in C_0\setminus A} \textsc{FS}(v,w)$\;
      \lIf{$|A|=k$}{
        \Return{$A$}
      }
    }
  
    \tcc{2. Select by overhead.}
    \While{$C_1\setminus A\not=\emptyset$}{
      $A \gets A+\text{argmin}_{n\in C_1\setminus A} \sum_{u, w\in N(v)} \textsc{OH}(uvw, A) \cdot \textsc{OH}{(uvw, A+n)}^2$\;
      \lIf{$|A|=k$}{
        \Return{$A$}
      }
    }
  
    \Return{$A$}
    \caption{Atomic strip selection}\label{alg:pcpp:atomicstripselection}
  \end{algorithm2e}
  
  If a waypoint $v\in P$ has a coverage value, i.e., $\text{val}(v)>0$, we still have to select the dominant strip that can only be skipped at the cost of the opportunity loss.
  For very straight passages, there may be no atomic strip that can be used without overhead (this can also be due to numeric issues).
  Thus, we propose a more dynamic approach.
  
  The selection of the dominant strip $a$ out of the set $A$ is performed by the \emph{usage} of the atomic strips, as defined in the following function:
  \begin{align*}
    &\textsc{SelectDominant}(v, A)=\\&\text{argmax}_{a\in A} \sum_{u,w \in N(v)} x_{uvw}\cdot \textsc{Usage}(uvw, a)
\end{align*}
  Let $\textsc{Turn}_a(u,v,w)$ be the turn angle if the passage $uvw$ is forced to use the atomic strip $a$.
  The usage depends on the turning overhead induced by forcing a passage to use it.
  If there is no overhead, the usage is $1.0$, and if there is overhead, the usage drops exponentially.
  \[\textsc{Usage}(uvw, a) = \lambda^{\left(\textsc{Turn}_a(u,v,w)-\textsc{Turn}(u,v,w)\right)/\phi}\]
  $\lambda$ is the usage at an additional turning angle of $\phi$, see \cref{tab:cpp:prac:alg:usage} for an example.
  Higher values allow more gap, which is necessary if the grid is not regular.
  We use $\lambda = 0.25$ and $\phi=\SI{45}{\degree}$ in the experiments.
  \begin{table}
    \centering
    \begin{tabular}{c c}
      \toprule
      Overhead & Usage \\
      \midrule 
      $\SI{0}{\degree}$ & $1.00$\\
      $\SI{1}{\degree}$ & $0.93$\\
      $\SI{2}{\degree}$ & $0.85$\\
      $\SI{5}{\degree}$ & $0.68$\\
      $\SI{10}{\degree}$ & $0.46$\\
      $\SI{30}{\degree}$ & $0.10$\\
      $\SI{50}{\degree}$ & $0.02$\\
      \bottomrule
    \end{tabular}
    \caption{Usage for $\lambda=0.1$ and $\phi=\SI{30}{\degree}$}\label{tab:cpp:prac:alg:usage}
  \end{table}
  
  An example for different $k$ can be seen in \cref{fig:pcpp:alg:atomicstripsexample}.

\section{Grids and Meshes}\label{sec:cpp:prac:grid}

\begin{figure}[htb]
  \centering
  \includegraphics[width=\columnwidth]{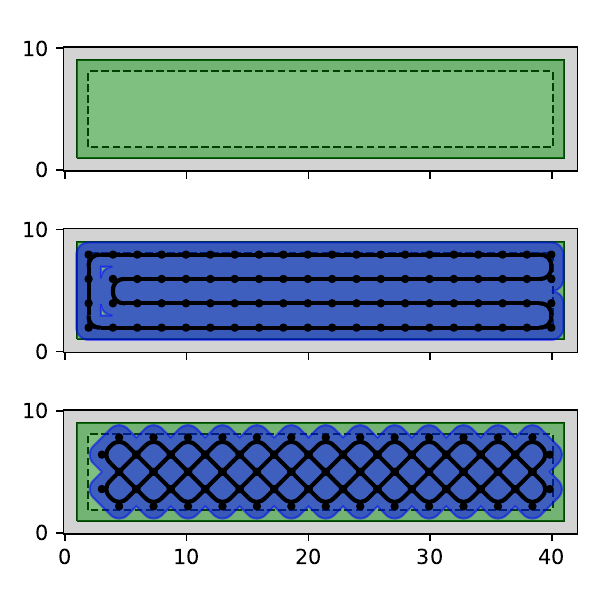}
  \caption{An example that allows arbitrary high turn costs for an unsuitable alignment.
  The upper image shows the area to be covered in green.
  The middle image shows a well-aligned grid, and the lower image an unsuitable grid that requires many turns.
  The blue area is the covered area, and the black lines show a tour computed by the algorithm.}\label{fig:grid:verybadgrid}
\end{figure}

An essential point for the grid-based methods is to choose a good grid.
Choosing an unsuitable grid, or even just the wrong orientation for it, can drastically reduce the achievable performance as also noted by Bormann et al.~\cite{bormann2018indoor}.
Especially in the presence of turn costs, orientating a square grid in \SI{45}{\degree} to a straight boundary will result in many turns, because all vertices at the boundary will need to make a turn.
Even if we would be able to compute an optimal tour for this grid, the gap to the optimal solution independent of the grid can be nearly arbitrarily large, see \cref{fig:grid:verybadgrid}.
Choosing a good grid, hence, is as important as finding a good solution within this grid.

In this section, we focus purely on how to convert the polygonal instance into a graph-based instance.
The reason for separating this from the algorithm evaluation is that this section can be applied to all grid-based coverage path planning algorithms.
While we use our algorithmic approach for evaluation, most observations can be transferred to general coverage path optimizations.
More specifically, we evaluate the following questions:

\begin{itemize}
    \item How well do regular square and triangular grids perform in terms of touring costs and coverage?
    \item What is the optimal edge length in terms of touring costs and coverage?
    \item How can we create good triangular and quadrilateral meshes to approximate the area?
    \item How do regular grids and meshes perform for full coverage and partial coverage?
\end{itemize}

We start with regular grids, as many algorithms only support regular grids, and then continue to irregular grids, i.e., meshes.

\subsection{Regular Grids}

Before we go to the lawless meshes, let us take a look at regular grids.

\begin{figure}
  \centering
  \includegraphics[width=\columnwidth]{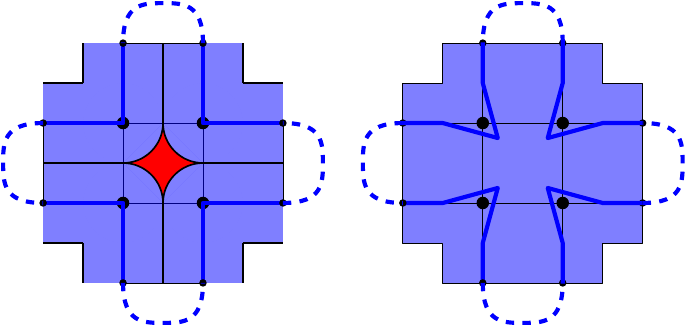}
  \caption{With a distance of $2r$, the middle of a square is not fully covered (red). However, we minimize turns such that these scenarios are also minimized. If these scenarios happen, they can be fixed with reasonably small costs by moving the turning points slightly to the inside. Depending on how the subtours are connected, changing the connections can also be useful to obtain smaller turns.}\label{fig:grid:square2rproblem}
\end{figure}

The most common grid is the square grid, which essentially partitions the area into small squares.
There are two options for the edge lengths, i.e., the distance between two adjacent vertices, for a circular tool with radius $r$: $2r$ or $\sqrt{2}r$.
The first option reflects the optimal distances between two parallel trajectories, while the second option already provides a full coverage by simply visiting all vertices.
In case of turns, a length of $2r$ will leave out a portion of the area, as can be seen in \cref{fig:grid:square2rproblem}.
Luckily, when minimizing turns, these cases are also minimized but not completely eliminated.
The remaining cases can be fixed by slightly moving the waypoints, as shown to the right, but we leave this technique to future work.

Another common grid is the triangular grid.
These are also known as hexagonal grids, because the dual graph consists of hexagons but the graph itself consists of triangles.
Here, every vertex has six neighbors, which makes its optimization more challenging but also allows more complex turns.
For achieving a full coverage by only visiting the vertices, we need a distance of $\frac{3}{\sqrt{3}}r$ to reach the center of each triangle.
If we want two parallel trajectories to be perfectly apart, we need a distance of $\frac{4}{\sqrt{3}}r$.
In this case, we again can lose coverage at turns, which we may need to fix.

\begin{figure}[pt]
  \centering
  \includegraphics[width=\columnwidth]{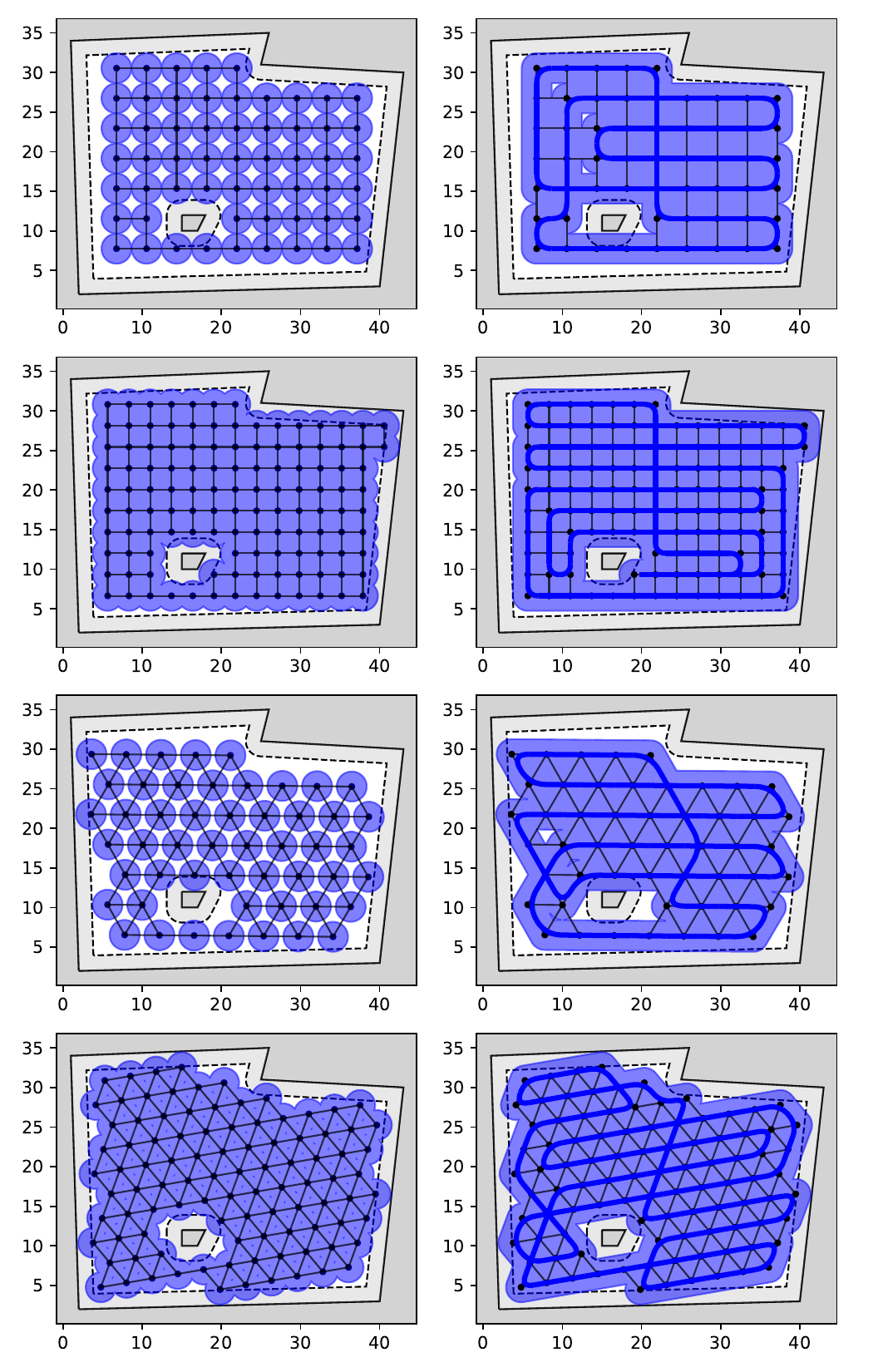}
  \caption{Square and triangular grids with point-based (denser) and line-based (sparser) distances.
  The blue areas on the left indicate the covered area from the vertices of the grid.
  The blue areas on the right indicate the covered area of a corresponding tour (blue trajectory).
  While the line-based grids do have an insufficient coverage from the vertices alone, the tour only leaves few gaps at turns but is much shorter.
  The white area enclosed by the dashed boundary describes the feasible tool positions.}\label{fig:grid:differentGridsExamples}
\end{figure}

For square and triangular grids, we denote the denser version that covers the interior area at the vertices as \emph{point-based}.
The other version, which can miss area at turns, but covers everything between two parallel lines, as \emph{line-based}. 
This results in four regular grids: \emph{point-based square grids}, \emph{line-based square grids}, \emph{point-based triangular grids}, and \emph{line-based triangular grids}.
Examples with covered area from the vertices and tours can be seen in \cref{fig:grid:differentGridsExamples}.
In those, we can also see that the greatest loss of coverage happens at the boundary.
The point-based grids have a small advantage here because they can place more points due to their finer resolution.
An easy solution to increase the coverage is to simply sweep once along the boundary, but this can be expensive for curvy boundaries.
We will look into other solutions with irregular grids later.

\begin{figure}[htb]
  \centering
  \includegraphics[width=\columnwidth]{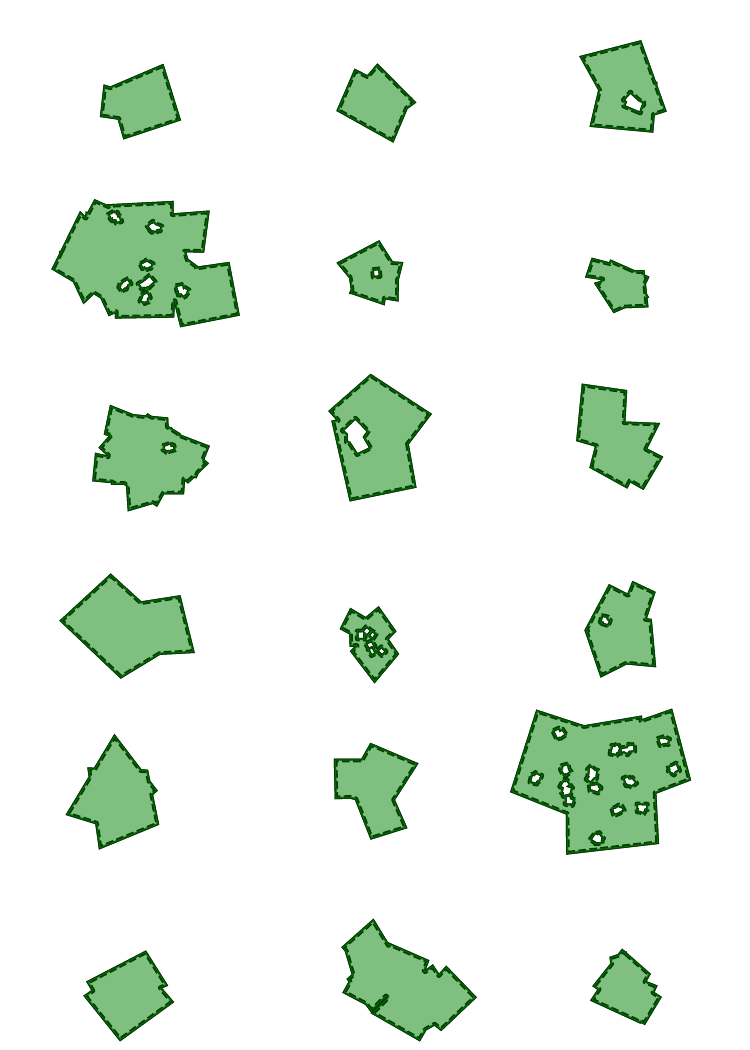}
  \caption{Examples of the used (full-coverage) instances.}\label{fig:grid:exampleGridExperiments}
\end{figure}
For comparing the different grids, we focus on polygons that can be reasonably well covered, i.e., that do not have narrow corners.
We do not consider (integral) orthogonal instances, inspired by simple rooms, as these can clearly be converted to a square grid and pose no serious challenge.
Instead, we create instances that are in the shape of more complicated architecture with many non-parallel lines.
Additionally, we add obstacles for some instances.
The instances are generated by merging multiple distorted rectangles and adding some holes with the same procedure.
During this process, only steps that do not lead to narrow corners, bottlenecks, or even disconnection are chosen.
By using a set of different random parameters for repetitions, sizes, and distortion strength, we generated a set of \num{200} instances.
Examples of these instances can be seen in \cref{fig:grid:exampleGridExperiments}.
The weight for the turn costs is $1.0,5.0, 50.0$ (measured in radian), and the tool-radius is uniformly set to $1.0$.
For simplicity, we focus on full coverage and only compare the touring costs and the coverage (aiming for \SI{100}{\percent}).
Partial coverage instances have many more parameters and are more difficult to compare.

Because an unsuitable alignment can make most grids very expensive, we try \num{20} random alignments of each grid, plus one which we rotate such that the sum of the minimal passage costs per vertex are minimal.
Of these, we choose the alignment with the minimal touring costs.

\begin{figure*}[tb]
  \centering
  \begin{subfigure}[b]{0.45\textwidth}
  \centering
  \includegraphics[width=\columnwidth]{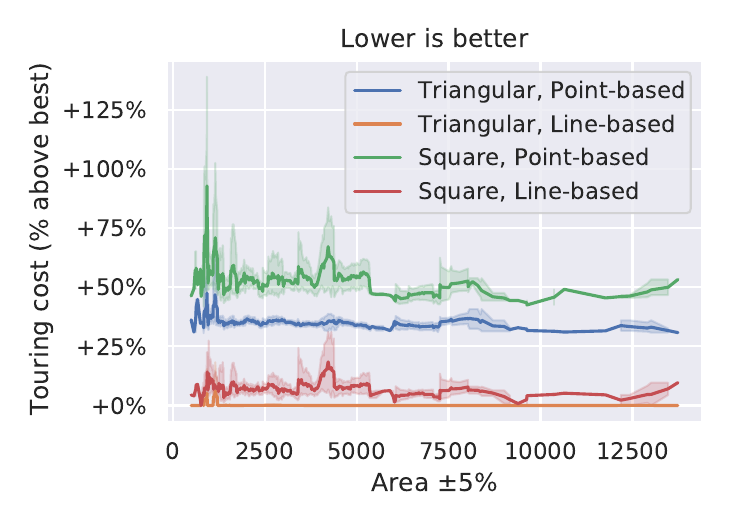}
    \caption{Area.}
  \end{subfigure}
  \begin{subfigure}[b]{0.45\textwidth}
  \centering
  \includegraphics[width=\columnwidth]{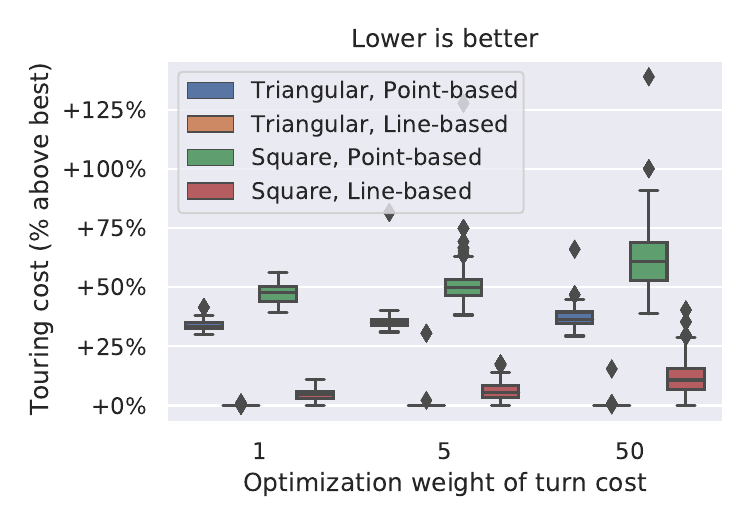}
  \caption{Turn costs.}
  \end{subfigure}
  \caption{%
    Touring costs for the different regular grids.
    To counter unsuitable alignments, only the best alignment is used.
    The value shows how many percent the touring costs are on average higher than the least expensive tour among the solutions for the corresponding instance.
    Line-based grids yield clearly less expensive tours, and this effect is stable also for larger instances.
    The triangular grid only has a small advantage over the square grid of less than \SI{7}{\percent}.
    The right plot (b) shows that square grids become worse for higher turn costs.}\label{fig:grid:differntGridsCosts}
\end{figure*}

Let us first take a look into the touring costs on each of the grids.
Of course, they can differ for different algorithms, but it still gives a good indication of the quality of the grid.
For comparing the touring costs of the different instances and solutions, we need to normalize the objectives.
This is done by putting the objective in relation to the best of the corresponding instances.
A value of \SI{50}{\percent} means that the tour is \SI{50}{\percent} more expensive than the least expensive solution.
The plots in \cref{fig:grid:differntGridsCosts} show that the \emph{line-based} grids result in the least expensive tours by a significant margin of over \SI{30}{\percent}.
The triangular grid has a small advantage of less than $\sim\SI{7}{\percent}$ over the square grid.
The point-based square grid performs worst and is on average \SI{50}{\percent} more expensive than the line-based triangular grid.
The results are relatively stable also for larger instances.
When comparing based on the turn cost weights, the triangular grids are very stable, while the square grids become slightly worse for higher turn costs.
This can be explained by the reduced turning abilities (only having \SI{90}{\degree} and \SI{180}{\degree}) of the square grid.

\begin{figure}
  \centering
  \includegraphics[width=0.7\columnwidth]{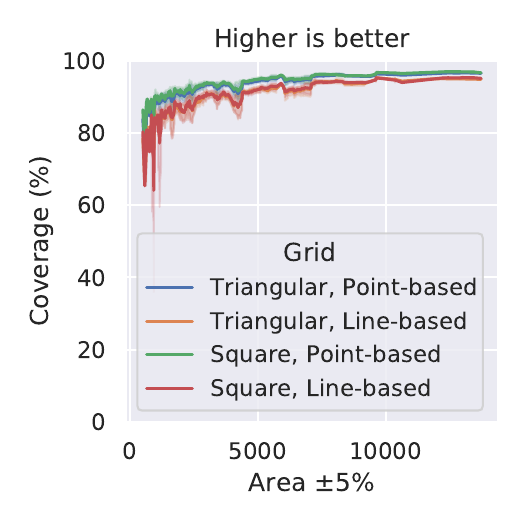}
  \caption{The covered area of the various grid types. Point-based instances only have a small advantage.}\label{fig:grid:coverageOfRegularGrids}
\end{figure}
Tours on line-based grids are, thus, less expensive than point-based grids, but how much does the coverage suffer?
If the coverage is too reduced, the lower touring costs are of little comfort.
Fortunately, the data in \cref{fig:grid:coverageOfRegularGrids} shows that the point-based triangular grid only covers on average less than \SI{4.2}{\percent} more than the line-based triangular grid.
Considering also that the point-based grids miss a few percent of the coverage even for the larger instances, this is sufficient for many applications.
The missed coverage that is especially high for smaller instances can be primarily attributed to the boundary.
For larger instances, the boundary ratio gets smaller and, thus, the coverage gets better for all grids.
In the end, we can decide between a \SI{4.2}{\percent} higher coverage using point-based grids, or a $\sim\SI{25}{\percent}$ less expensive tour.

\begin{figure}[htb]
  \centering
  \includegraphics[width=0.6\columnwidth]{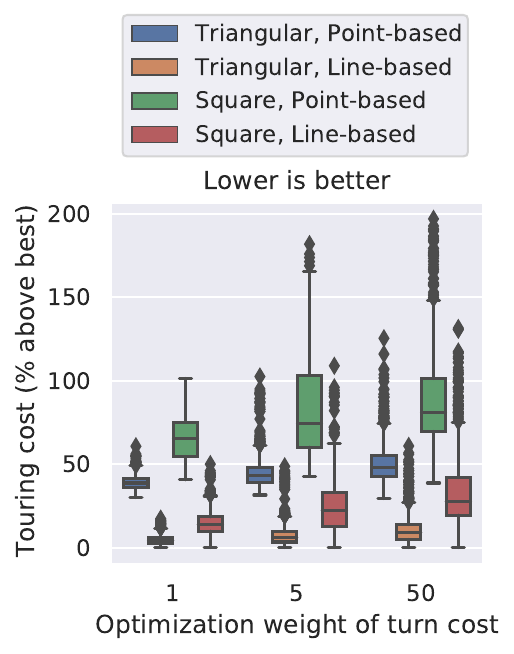}
  \caption{Average touring costs for the random alignment compared to the best solution.}\label{fig:cpp:prac:grid:avgalignment}
\end{figure}

We selected the best alignment out of \num{20} random alignments and one optimized alignment, which is, of course, a deceptive selection.
As optimizing a tour in a grid takes some time, simply trying many random alignments is not very efficient.
The natural question that comes to mind is, how well the average random alignment performs.
This question can be answered by the plot in \cref{fig:cpp:prac:grid:avgalignment}: for some instances not very well.
The higher the weight of the turn costs, the worse is the average alignment.
While for many instances, the random alignments of the line-based triangular grids are still reasonably good, there is a high deviation with many outliers.
For the square grids, the random alignment is most of the time unsuitably aligned, implying that it needs a more careful alignment than triangular grids.

\begin{figure*}[tbp]
  \centering
  \includegraphics[width=0.8\textwidth]{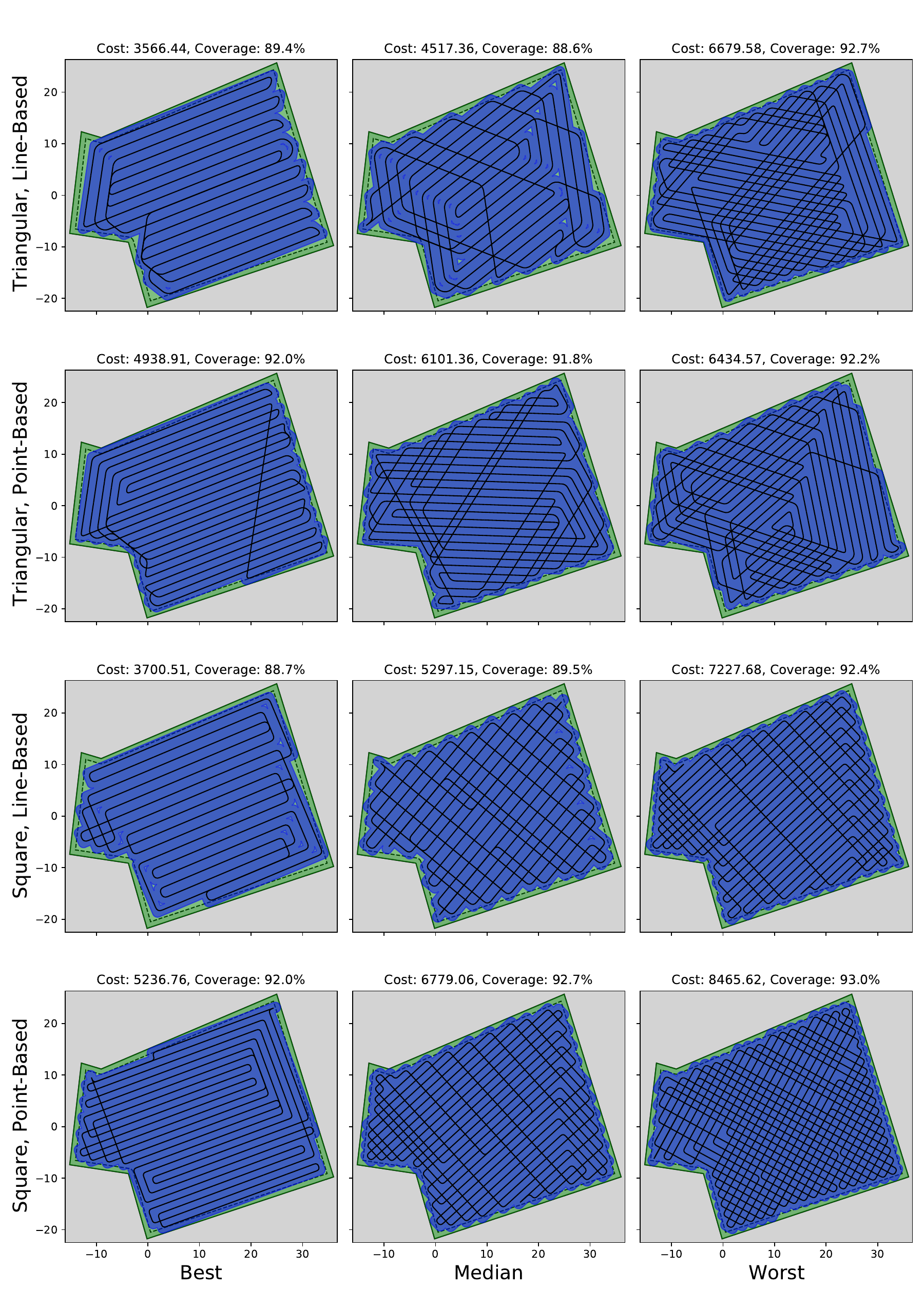}
  \caption{Examples for tours on the four grid types with different alignments. In the first column, the best alignment is shown. In the middle, an alignment that yielded an average tour; and on the right, the worst alignment for the grid type on this instance. The high weight on the turn costs produces lots of redundant straight coverages if the boundary is badly aligned to the grid.}\label{fig:grid:differentGridsMinMaxMean}
\end{figure*}

Examples for good, bad, and median alignments can be seen in \cref{fig:grid:differentGridsMinMaxMean}.
The selection only considers the touring costs.
The concrete instance has a high turn cost weight of \num{50}, which leads to the high redundancy for the worse alignments, as the tour tries to only make turns at corner points where it has to make turns anyway.

For the experiments in the following sections, regular grids will be line-based triangular grids oriented by our heuristic.
This heuristic, as also explained above, rotates the grid such that the sum of minimal passing costs for all vertices is minimal.
The idea behind this is that we minimize the number of `staircase'-boundaries that can be observed in the bad alignments in \cref{fig:grid:differentGridsMinMaxMean}.
This heuristic performs reasonably well, and is on average only \SI{2.7}{\percent} more expensive than the best grid in the previous experiment.
In comparison, the best of \num{20} random grids was only \SI{0.6}{\percent} more expensive, still showing some room for improvement for our heuristic.
For the other grid types, the best of \num{20} random grids was also slightly better than our heuristic but only minimally (e.g., \SI{10}{\percent} vs. \SI{9.2}{\percent} for line-based square grids.
).
Overall, the heuristic works reasonably well and is much faster than computing the solution on \num{20} random alignments and returning the best solution.

\subsection{Meshes}

We have seen how to approximate a polygonal area using a square or triangular grid.
These regular grids are much easier to work with and are therefore very common for coverage path planning.
However, the approximation can be very crude, not only missing large parts of the boundary but also being badly aligned to it.
Luckily, we are not the only ones with the desire to approximate an area by a grid-like structure, and we can make use of the many results in the field of mesh generation.
In this section, we consider the use of meshes instead of regular grid graphs.

The field of \emph{mesh generation} (also known as grid generation, meshing, or gridding) considers the partition of a surface or geometric object into simpler elements, such as triangles or quadrilaterals (or three-dimensional counterparts).
This is a fundamental task in computer graphics, physics simulation, geography, and cartography to deal with the more complex objects.
The concrete specifications of a good mesh differ for some tasks, but import properties often are:
\begin{itemize}
  \item Angles should not be too small or too wide. A primary motivation for this is due to numerical issues.
  \item The mesh should have a low complexity, i.e., minimize the number of introduced vertices and edges.
  \item The mesh should be as regular as possible, i.e., low variance in size and shape.
    Most algorithms allow the user to specify a desired or maximal edge length.
\end{itemize}
To find out more about the general subject of mesh generation, you can take a look into the surveys~\cite{BernP00,Bern04} that are part of Computational Geometry books.

\begin{figure}
  \includegraphics[width=\columnwidth]{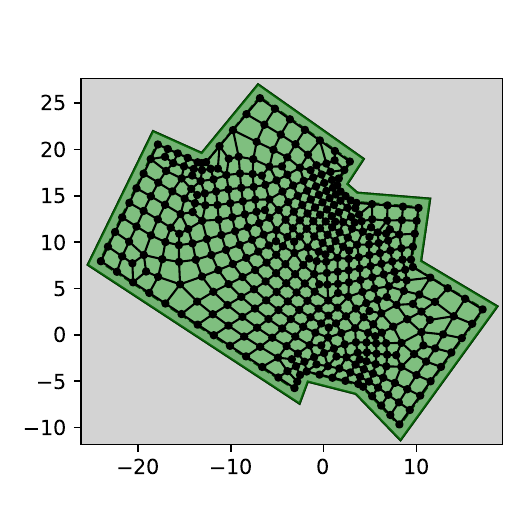}
  \caption{A quadrilateral mesh that is sensitive to the close points on the boundary and reacts with a very dense grid in these areas. Generated with \emph{gmsh}~\cite{geuzaine2009gmsh} using the \emph{Frontal-Delaunay for Quads} algorithm with recombination.}\label{fig:cpp:prac:grid:baddensity}
\end{figure}

\begin{figure}[tb]
  \begin{subfigure}[b]{0.24\columnwidth}
    \includegraphics[width=\columnwidth]{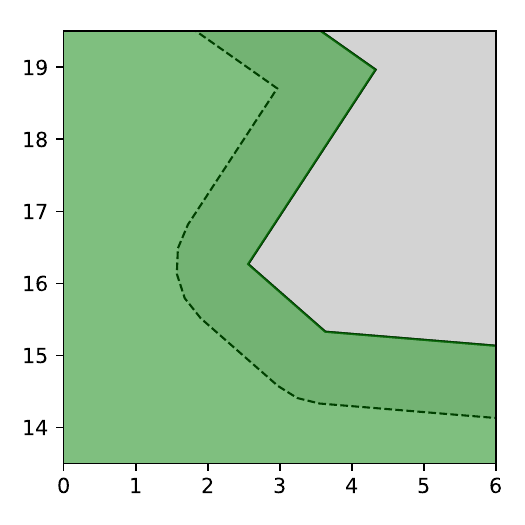}
    \caption{}
  \end{subfigure}
  \begin{subfigure}[b]{0.24\columnwidth}
    \includegraphics[width=\columnwidth]{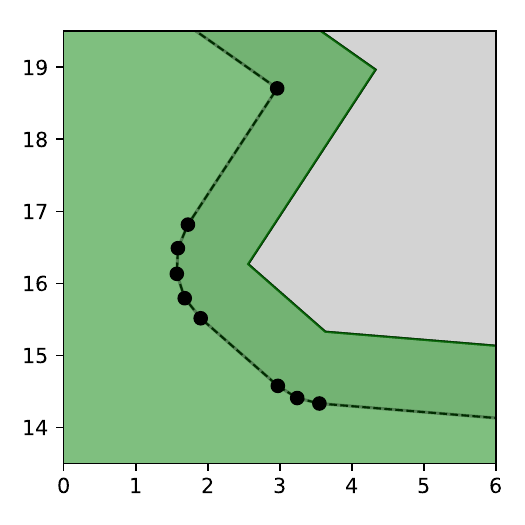}
    \caption{}
  \end{subfigure}
  \begin{subfigure}[b]{0.24\columnwidth}
    \includegraphics[width=\columnwidth]{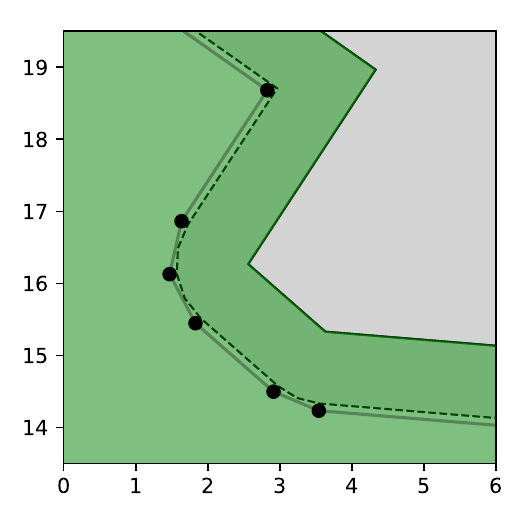}
    \caption{}
  \end{subfigure}
  \begin{subfigure}[b]{0.24\columnwidth}
    \includegraphics[width=\columnwidth]{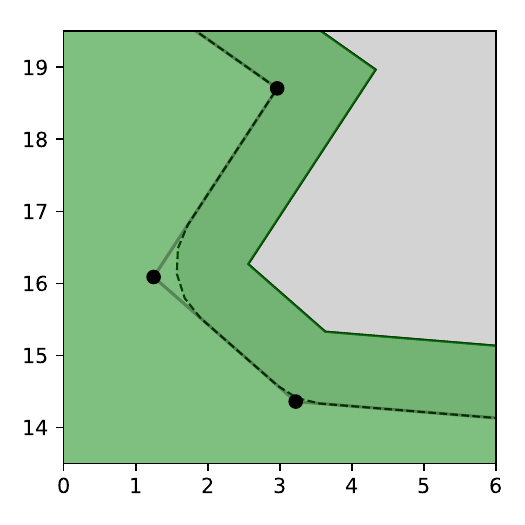}
    \caption{}
  \end{subfigure}
  \caption{Concave corners (a) in the area result in circular curves, under the assumption of a circular robot.
          Approximating such a curve results in many close boundary points (b) that are problematic for many mesh generation methods.
          To circumvent this problem, we should simplify the resulting boundary (c) by using, e.g., the Douglas-Peucker algorithm~\cite{douglas1973algorithms}.
          Because this can result in intersecting the boundary, the points should be moved slightly inwards.
          Additionally, we can work on a shrank boundary polygon (d) instead of the feasible area (which potentially leaves a small part of the area uncovered).}\label{fig:cpp:grid:simplify}
\end{figure}

Many mesh generators are very sensitive to close points on the boundary and react by placing equally close points that result in a high density in this area, see \cref{fig:cpp:prac:grid:baddensity}.
This high density is, of course, bad for coverage path planning and yields tours with much redundant coverage.
Close points on the boundary are frequently created by concave corners in the room, around which a circular robot has to make a circular turn, approximated by polygon with many segments.
There are two steps we should perform, as shown in \cref{fig:cpp:grid:simplify}:
First, we can simplify the polygon and remove close points, e.g., by the usage of the Douglas-Peucker algorithm~\cite{douglas1973algorithms}.
Because the connecting lines at the boundary can now intersect the unreachable boundary area, we have to move the points slightly inwards.
Second, we can simply shrink the boundary by the robot radius, instead of computing the feasible area by a Minkowski sum.
The resulting coverage path may miss some small area at this corner, but the underlying mesh allows a much better tour.
We still need to perform a simplification for complicated boundary parts not created by curve approximations.

\begin{figure}
  \begin{subfigure}[b]{0.49\columnwidth}
    \includegraphics[width=\columnwidth]{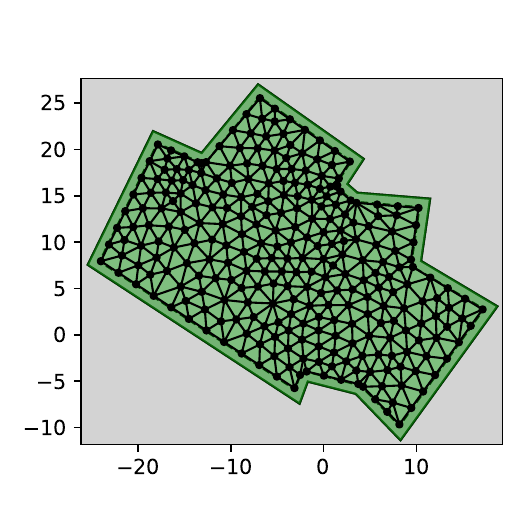}
    \caption{Without smoothing.}
  \end{subfigure}
  \begin{subfigure}[b]{0.49\columnwidth}
    \includegraphics[width=\columnwidth]{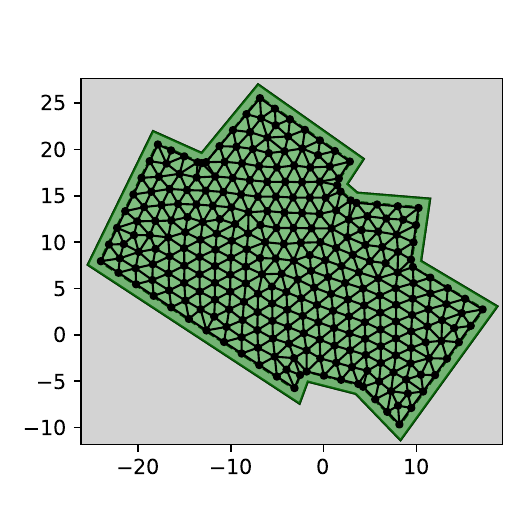}
    \caption{With smoothing.}
  \end{subfigure}
  \caption{Many mesh generators only yield good meshes after smoothing.}\label{fig:cpp:prac:grid:smoothing}
\end{figure}

One does not need smooth meshes for all applications, and therefore many mesh generators only yield very rough grids out of the box.
In combination with smoothing and optimization methods, these generators can still provide us with smooth meshes, see \cref{fig:cpp:prac:grid:smoothing}.
The classical algorithm of Du et al.~\cite{du1999centroidal} tries to achieve a centroidal Voronoi tessellation (CVT) by moving the points to the centroid of their Voronoi-cells (Lloyd's method~\cite{lloyd1982least}).
Instead of the Voronoi diagram, one can also work on the dual Delaunay graph, as proposed by Chen and Holst~\cite{chen2011efficient}.
The \emph{optimesh}-library~\cite{nico_schlomer_2021_4728056} implements these algorithms as well as some variants.
The corresponding GitHub-page (\url{https://github.com/nschloe/optimesh}) gives a great overview with animations and also some experimental analyzes.
Based on this experimental analysis, verified on some instances, we use the implemented CVT-variant \emph{cvt-full} for all triangular meshes with a tolerance of \num{1e-5} and \num{1000} iterations.

To compare all mesh generators is beyond the scope of this thesis, therefore we focus on a small selection of seven algorithms that seemed promising based on documentation and samples:
\begin{itemize}
  \item The \emph{MeshAdapt} ($\Delta$MA), \emph{Frontal-Delaunay} ($\Delta$FD), \emph{Frontal-Delaunay for Quads} ($\Delta$FDQ), and \emph{Packing of Parallelograms} ($\Delta$PP) methods of \emph{gmsh}~\cite{geuzaine2009gmsh} for triangular meshes.
  \item The \emph{Frontal-Delaunay for Quads} ($\square$FDQ), and \emph{Packing of Parallelograms} ($\square$PP) methods with recombination of \emph{gmsh}~\cite{geuzaine2009gmsh} for quadrilateral meshes.
  \item The \emph{dmsh}-algorithm~\cite{nico_schlomer_2021_5019221}
   that is inspired by \emph{distmesh}~\cite{persson2004simple}.
\end{itemize}
To apply the same conditions to all algorithms, we use the same polygon shrinking and simplification for all mesh generators, even if they can handle complex, curved boundaries reasonably well, such as \emph{dmsh}.

\begin{figure*}[tb]
  \begin{subfigure}[b]{0.3\textwidth}
    \includegraphics[width=\columnwidth]{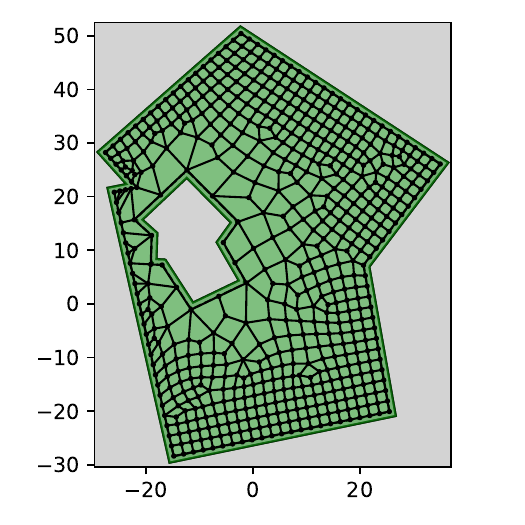}
    \caption{gmsh has problems with holes.}
  \end{subfigure}
  \begin{subfigure}[b]{0.3\textwidth}
    \includegraphics[width=\columnwidth]{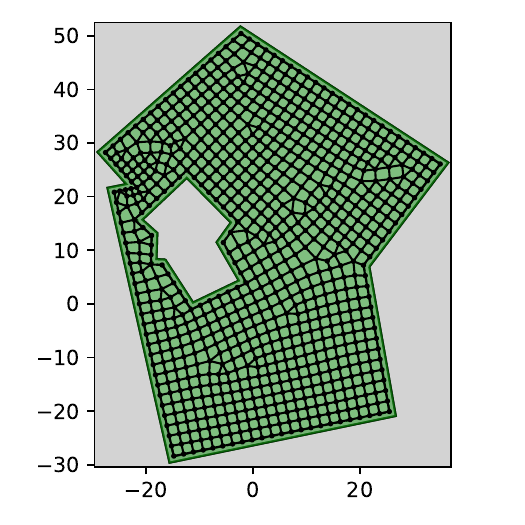}
    \caption{Fixed with workaround.}
  \end{subfigure}
  \begin{subfigure}[b]{0.3\textwidth}
    \includegraphics[width=\columnwidth]{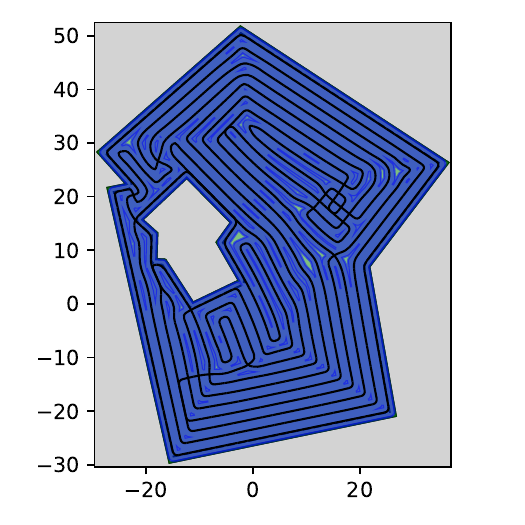}
    \caption{Resulting tour.}
  \end{subfigure}
  \caption{The meshing algorithms of gmsh often have problems with holes in the polygon (at least using the pygmsh interface~\cite{nico_schlomer_2021_5196231}). The boundary of the hole and the area surrounding it does not get populated enough. This can be fixed by enforcing fixed points on the hole boundary by replacing long segments with shorter ones. Using this fix, we can compute a good tour despite some artifacts in the mesh. This example uses \emph{Packing of Parallelograms} with recombination.}\label{fig:cpp:prac:grid:holeproblem}
\end{figure*}

The algorithms of \emph{gmsh} often have problems with holes, see \cref{fig:cpp:prac:grid:holeproblem}.
To fix this problem, we replace long segments in holes with shorter ones that approximate the desired edge length.
If $p_0p_1$ is a long segment and $d$ is the desired edge length with $d<||p_0-p_1||$, we replace $p_0p_1$ with $\approx \nicefrac{||p_0-p_1||}{d}+1$ equal sub-segments.

\begin{figure*}[tb]
  \begin{subfigure}[b]{0.45\textwidth}
    \includegraphics[width=\columnwidth]{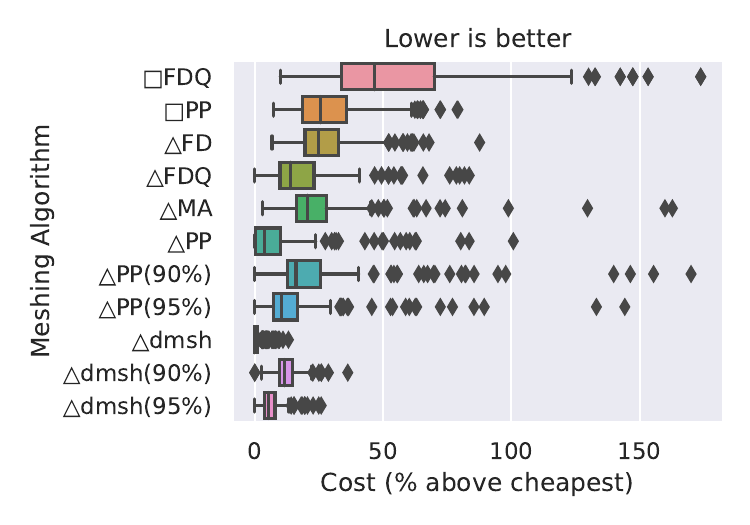}
    \caption{Touring costs (lower is better).}
  \end{subfigure}
  \begin{subfigure}[b]{0.45\textwidth}
    \includegraphics[width=\columnwidth]{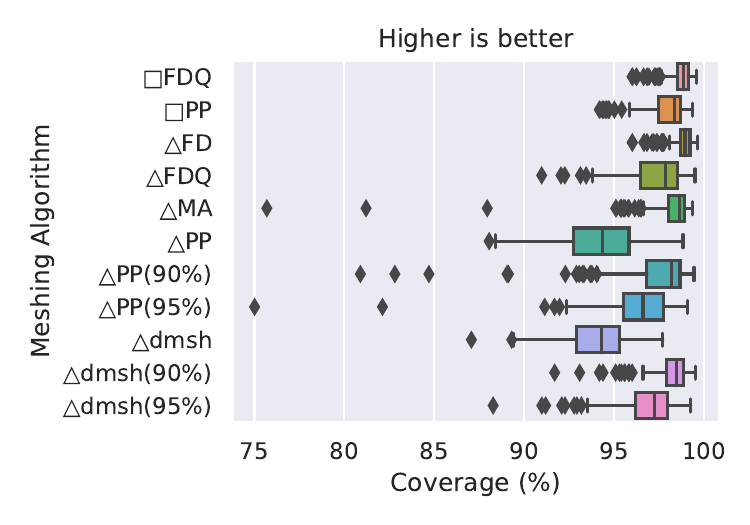}
    \caption{Coverage (higher is better).}
  \end{subfigure}
  \caption{Cost and coverage of tours in various meshes.
  The $\square$ and $\Delta$ indicate quadrilateral resp.\ triangular meshes.
  We can see low costs for \emph{dmsh} and triangular \emph{Packing of Parallelograms}, and a good coverage for the \emph{Frontal-Delaunay}.}\label{fig:cpp:prac:grid:mesh:results}
\end{figure*}

We compare the meshes on the same full coverage instances as the regular grids (\cref{fig:grid:exampleGridExperiments}).
The focus on full coverage allows us to focus more attention onto the coverage quality and the density of the meshes.
The plots in \cref{fig:cpp:prac:grid:mesh:results} show the average touring costs and coverage of the different meshes.
The touring cost is measured in how much more expensive a tour is compared to the least expensive tour we found for this instance.
A value of $\SI{20}{\percent}$ would indicate that the mesh yields tours that are on average $\SI{20}{\percent}$ more expensive than the least expensive tour.
The coverage is measured regarding the whole area (note that not all of the area is always reachable).
A value of, e.g., $\SI{95}{\percent}$ would indicate that the mesh yields tours that cover on average $\SI{95}{\percent}$ of the whole area.

Area is left uncovered not only due to turns and boundaries (as for regular grids) but also because the meshes do not always match the desired distances and can be too sparse in some areas.
This especially happens for \emph{dmsh} and \emph{Packing of Parallelograms}, thus, we also reduced their mesh size by \SI{90}{\percent} and \SI{95}{\percent} to increase the coverage for these cases.
\emph{dmsh} failed to create a grid in \num{5} of the \num{200} instances because the internal geometric routines threw exceptions, likely due to numerical issues.
These results are therefore ignored in analysis.
One additional instance resulted in some disconnected areas after the polygon processing and is likewise ignored for all meshes.

The \emph{dmsh} and \emph{Packing of Parallelograms}-algorithms yield the least expensive tours on average, with \emph{dmsh} having an advantage for higher turn cost ratios.
However, when taking a closer look into the data, most meshes can achieve good grids for specific instances.
For example, relatively rectangular instances work very well with the quadrilateral mesh of \emph{Packing of Parallelograms}.
The coverage quality is nearly inverse to the touring costs.
The triangular \emph{Frontal-Delaunay} mesh performs best in coverage, and is on average around \SI{25}{\percent} more expensive than the least expensive tour (that potentially has a lower coverage).
Therefore, reducing the desired edge length for \emph{dmsh} and \emph{Packing of Parallelograms} results in better coverage at a small cost increase.

The mean runtime, as shown in \cref{tab:cpp:prac:mesh:runtime}, for some meshes can differ quite a lot with a mean runtime of \SI{146.2}{\second} for $\Delta$dmsh(90\%) and only \SI{89.5}{\second} for $\square$PP\@.
Interestingly, the quadrilateral meshes are not necessarily faster despite much smaller auxiliary graphs.
This is likely due to the overhead induced by inefficient implementations of auxiliary steps, while the already optimized steps of the linear program and the matching algorithm are expected to be faster.
Meshes created by \emph{dmsh} are relatively slow but, contrary to the \emph{gmsh}-algorithms, it is written in pure Python and, thus, not optimized for runtime.
Otherwise, the triangular \emph{Packing of Parallelograms} meshes are relatively fast.
Note that the runtime refers to the whole solution process and not just the grid generation.
The square grids are usually generated within one second and for the triangular grids, the optimization process of \emph{optimesh} (in Python with NumPy and SciPy) can take up to a few seconds.
\begin{table}
  \centering
  \begin{tabular}{l c}
    \toprule
    Mesh & Runtime (s)\\
    \midrule
$\square$FDQ & 129.7 \\
$\square$PP & 89.5 \\
$\Delta$FD & 137.6 \\
$\Delta$FDQ & 113.1 \\
$\Delta$MA & 120.7 \\
$\Delta$PP & 94.1 \\
$\Delta$PP(90\%) & 127.4 \\
$\Delta$PP(95\%) & 106.6 \\
$\Delta$dmsh & 114.6 \\
$\Delta$dmsh(90\%) & 146.2 \\
$\Delta$dmsh(95\%) & 123.3 \\
    \bottomrule
  \end{tabular}
  \caption{Mean runtime with different meshes.}\label{tab:cpp:prac:mesh:runtime}
\end{table}

\begin{figure*}[t]
  \centering
  \includegraphics[width=\textwidth]{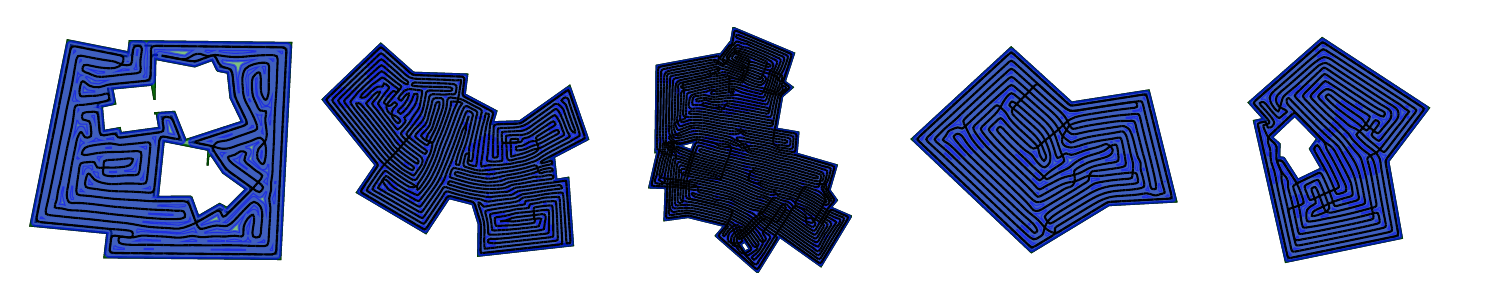}
  \caption{Only \emph{Packing of Parallelograms} yielded good quadrilateral meshes, shown here. The trajectory is displayed in black, the covered area in blue.}\label{fig:cpp:prac:mesh:examples:ppsquare}
\end{figure*}

\begin{figure*}[pt]
  \begin{subfigure}[b]{\textwidth}
    \includegraphics[width=\columnwidth]{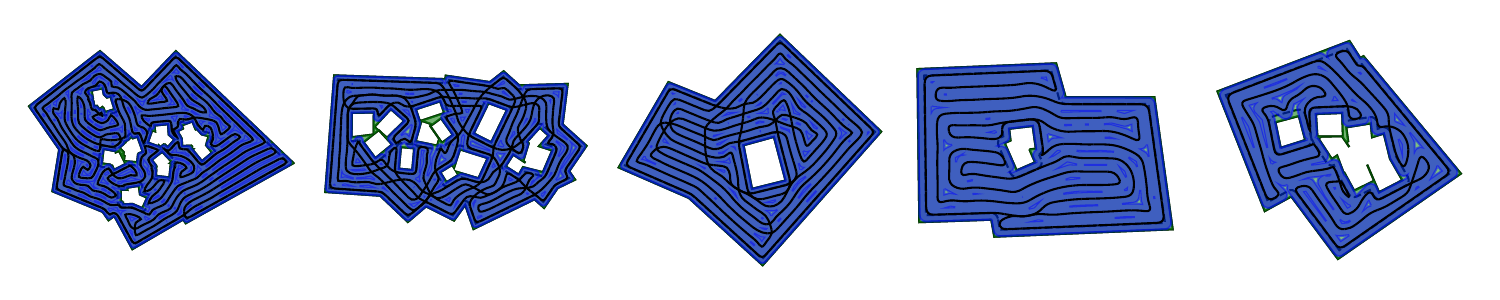}
    \caption{\emph{MeshAdapt} ($\Delta$)}
  \end{subfigure}
  \begin{subfigure}[b]{\textwidth}
    \includegraphics[width=\columnwidth]{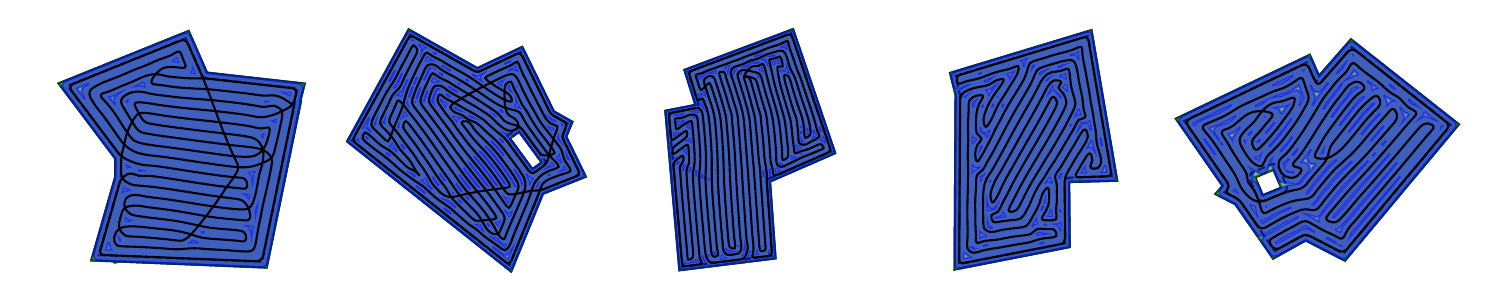}
    \caption{\emph{Frontal-Delaunay} ($\Delta$)}
  \end{subfigure}
  \begin{subfigure}[b]{\textwidth}
    \includegraphics[width=\columnwidth]{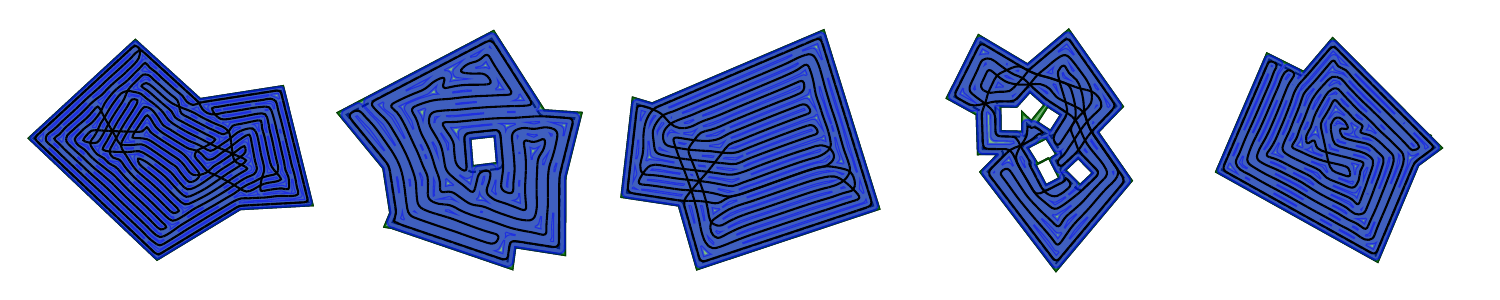}
    \caption{\emph{Frontal-Delaunay for Quads} ($\Delta$)}
  \end{subfigure}
  \begin{subfigure}[b]{\textwidth}
    \includegraphics[width=\columnwidth]{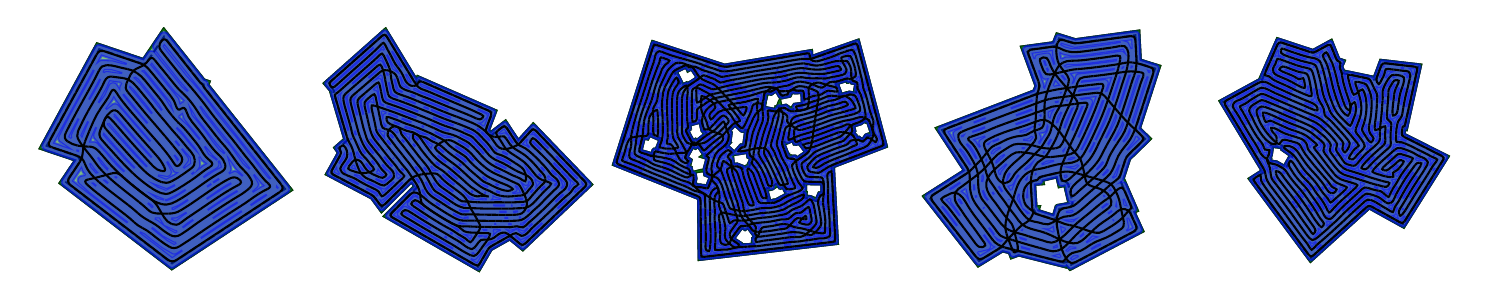}
    \caption{\emph{Packing of Parallelograms} ($\Delta$)}
  \end{subfigure}
  \begin{subfigure}[b]{\textwidth}
    \includegraphics[width=\columnwidth]{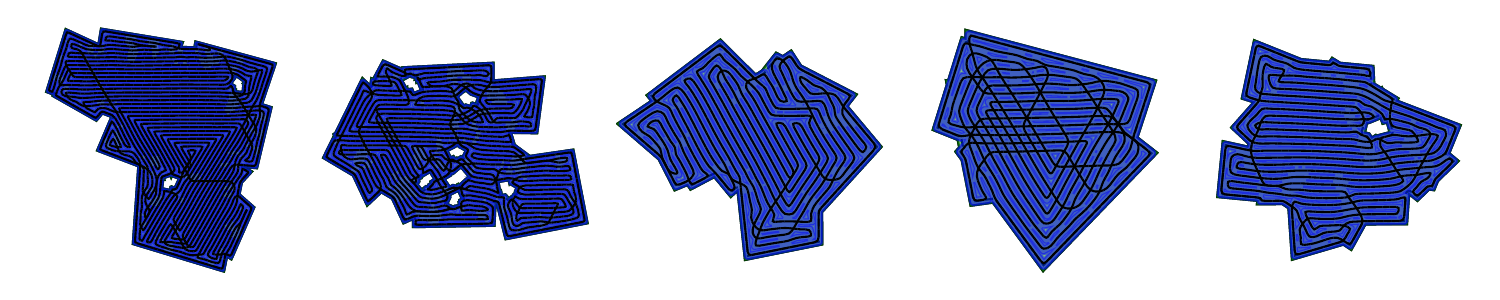}
    \caption{\emph{dmsh} ($\Delta$)}
  \end{subfigure}
  \caption{%
    Instances and tours with at least \SI{95}{\percent} coverage on which the corresponding mesh performed well (within \SI{5}{\percent} of the best).
    \emph{MeshAdapt} performs well on instances with large hole areas, while the two \emph{Frontal-Delaunay} versions perform primarily well for rather simple instances.
    By reducing the edge length in \emph{dmsh}, we improve the coverage while only slightly increasing the costs.
    We consider \emph{dmsh} with \SI{95}{\percent} edge length as our favorite, as it achieves a good performance and a sufficient coverage on average.
    If a more thorough coverage is necessary, we recommend simply to reduce the edge length of \emph{dmsh}.}\label{fig:cpp:prac:mesh:examples}
\end{figure*}

When looking onto which instances have been solved well on which mesh in \cref{fig:cpp:prac:mesh:examples,fig:cpp:prac:mesh:examples:ppsquare}, no clear pattern is visible.
The only observations that can be made are that \emph{MeshAdapt} performs well on instances with many or larger holes, and the \emph{Frontal-Delaunay} seems only to perform well on simpler instances.
In the next sections, we will use \emph{dmsh} with a \SI{95}{\percent} point distance.
In case of numerical issues, we will fall back on \emph{Packing of Parallelograms}, also with \SI{95}{\percent} point distance.
The coverage of these meshes is on average sufficiently high with \SI{96.9}{\percent} resp. \SI{96.3}{\percent}.

\subsection{Comparison}\label{sec:pcpp:grid:comparison}

We now compare the performance of meshes and regular grids for partial and full coverage.
For this, we generate instances as before based on the union of not-too-narrow quadrilaterals.
We do the same for valuable and expensive areas.
For the expensive areas, we combine all overlapping areas into a single area and do not allow overlap.
Otherwise, we can easily get extremely high factors.
The valuable areas are not multiplicative but additive and, thus, do not present this problem.
The unsteadiness of overlapping valuable areas can even make the instances more realistic.
Finding the right set of parameters such that we get well-balanced instances is difficult.
Using purely random parameters, one often ends up with full coverage because the values are too high, or no coverage because the touring costs are too high.
Thus, a careful fine-tuning of parameters is necessary to obtain diverse but interesting instances.
See \cref{fig:cpp:prac:grid:partialexamples} for a set of example instances with tours.

We evaluate the quality of the tours based on the objective, as described in \cref{eq:pcpp:obj}, which combines touring costs and coverage value.
To make the objectives comparable over all instances, we divide every objective by the minimal objective known for the corresponding instance.
We use the \emph{dmsh}-algorithm with \SI{95}{\percent} edge length as representative of mesh-based coverage, and a regular triangular grid with line-based distances for grid-based coverage.
To directly show the advantage of the partial-coverage technique, we also compare the approaches with enforced full coverage.

\begin{figure}[htb]
  \includegraphics[width=\columnwidth]{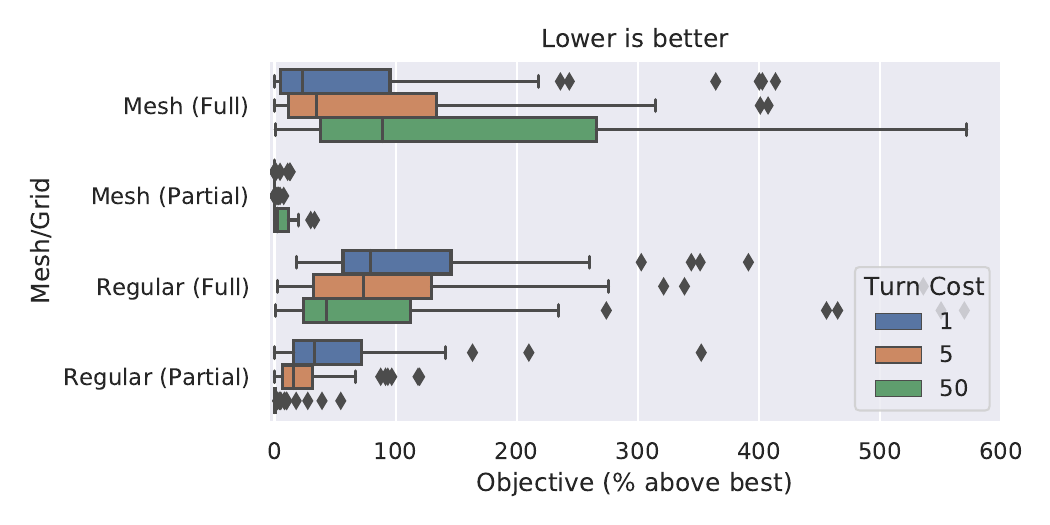}
  \caption{Relative objectives for partial coverage computed with meshes and regular grids.
  Additionally, we compare partial with full coverage.
  The results are split for different turning cost weights, with \num{50} having a higher focus on turns than the traveled distance.
  We can see that the ability of partial coverage has a significant advantage.
  The pure touring costs of the partial-coverage tours on the mesh as well as the regular grid are similar, but the coverage on the mesh is much better.}\label{fig:cpp:prac:grid:partialcomparison}
\end{figure}
In \cref{fig:cpp:prac:grid:partialcomparison}, we can see that
\begin{itemize}
    \item The partial mesh approach achieves the best objectives by a clear margin, independent of the weight of the turn costs.
      A deeper look into the data shows that the touring costs of the mesh are slightly worse than of the regular grid, but this method covers much more valuable area.
    \item The performance of the meshes drops for higher turn costs, while the performance of the regular grids improves.
    \item The full-coverage tours are much more expensive than the partial-coverage tours.
      Using only partial-coverage for the evaluated instances provides a significant advantage.
      Note that some drastic outliers have been removed from the plot.
      They can be explained by small valuable areas that can be covered by a small tour which is obviously better than a full-coverage tour.
      However, many instances have larger valuable areas, and even the partial tours often cover much of the area.
\end{itemize}

\begin{figure}
\begin{subfigure}[b]{0.49\columnwidth}
  \includegraphics[width=\columnwidth]{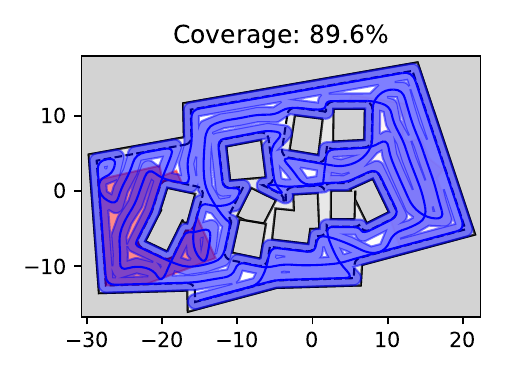}
  \caption{Mesh}
\end{subfigure}
\begin{subfigure}[b]{0.49\columnwidth}
  \includegraphics[width=\columnwidth]{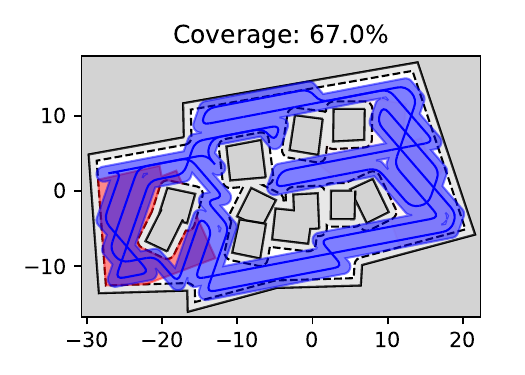}
  \caption{Regular Grid}
\end{subfigure}
\caption{An example for a difficult instance for which a mesh still achieves a reasonable coverage, but a regular grid does not.
A regular grid is too rigid to fit into the narrow passages.}\label{fig:cpp:prac:grid:badcoverage}
\end{figure}

\begin{figure}[htb!]
  \begin{subfigure}[b]{0.49\columnwidth}
    \includegraphics[width=\columnwidth]{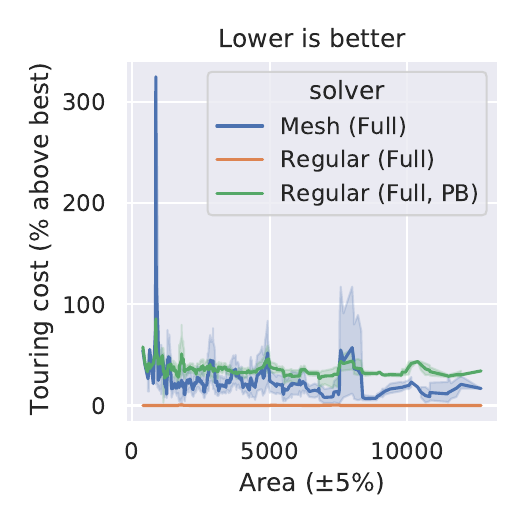}
  \end{subfigure}
  \begin{subfigure}[b]{0.49\columnwidth}
    \includegraphics[width=\columnwidth]{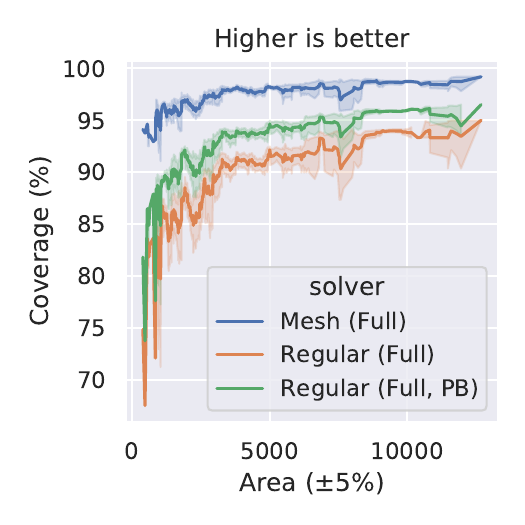}
  \end{subfigure}
  \caption{Meshes often yield slightly more expensive tours for full coverage, but also achieve a much better coverage.
  When using point-based (PB) edge lengths, the regular grid performs worse for full and partial coverage.
  The coverage generally increases for larger instances.
  The outliers are often small and complicated instances that a regular grid cannot cover properly but a mesh can.
  This results in high touring costs for the mesh and a low coverage for the regular grid.}\label{fig:cpp:prac:grid:fullcomp}
\end{figure}

When considering only the touring costs for full coverage, regular grids show to yield less expensive tours but with a lower coverage, as can be seen in \cref{fig:cpp:prac:grid:fullcomp}.
Even when using point-based edge lengths for the regular grids, the coverage cannot compete with meshes while the costs are even higher.
The significantly worse coverages often appear in complicated, small instances like in \cref{fig:cpp:prac:grid:badcoverage}.
A regular grid cannot fit to the complex and narrow environment while a mesh can.
Such instances also result in much higher touring costs for meshes because they do not skip the difficult areas.
However, the meshes also leave more area uncovered than in less narrow areas.
Because the mesh only tries to fit an average distance, the density in narrow areas can become too low, resulting in gaps.

For the further experiments, we focus on meshes because they achieve a more reliable coverage even for difficult instances.

\end{document}